%% file: main.tex
\documentclass[letterpaper]{article}
\usepackage[preprint]{main}  
\usepackage[hyphens]{url} 
\usepackage{graphicx} 
\usepackage{natbib}  
\usepackage{caption} 
\usepackage{algorithm}
\usepackage{algorithmic}
\usepackage{amsmath}
\usepackage{amssymb}

\usepackage{newfloat}
\usepackage{listings}
\DeclareCaptionStyle{ruled}{labelfont=normalfont,labelsep=colon,strut=off}
\floatstyle{ruled}
\newfloat{listing}{tb}{lst}{}
\floatname{listing}{Listing}

\usepackage{booktabs}
\usepackage[table]{xcolor}
\usepackage{makecell}
\usepackage{array}
\usepackage{tabularx}

\newcolumntype{L}[1]{>{\raggedright\arraybackslash}p{#1}}
\newcolumntype{Y}{>{\centering\arraybackslash}X}

\title{Reduce Tokens, Refine Semantics: Training-Free Acceleration of\\Vision-Language Models}

\author{
    Jiasheng Li\textsuperscript{\rm 1},
    Zhong Ji\textsuperscript{\rm 1}\corresponding,
    Yan Zhang\textsuperscript{\rm 2},
    Huihui Li\textsuperscript{\rm 1}
}

\affiliations{
    \textsuperscript{\rm 1}Tianjin University, Tianjin, China\\
    \textsuperscript{\rm 2}Tsinghua University, Beijing, China\\
    li\_jiasheng@tju.edu.cn,
    jizhong@tju.edu.cn,
    yzhang1995@tsinghua.edu.cn,
    huihuili@tju.edu.cn
}

\begin{document}

\maketitle

\begin{abstract}
Long visual token sequences impose substantial inference costs on large vision-language models (VLMs), motivating methods that reduce visual tokens while preserving semantic fidelity. Existing methods primarily focus on pruning or compressing redundant tokens without sufficiently examining whether the resulting compact representations remain semantically consistent with the full visual input. Reducing an $N$-token sequence to $K$ tokens can discard, dilute, or misassign critical visual cues, causing semantic drift that compromises visual understanding. We introduce \textbf{RefineV}, a training-free framework that couples token reduction with semantic refinement by calibrating compact visual representations before reasoning. RefineV comprises two complementary modules. 
Perturbation-Robust Calibration Anchoring selects compact anchors with stable model-side influence across multiple perturbation directions. Confidence-Gated Token Calibration then screens residual tokens according to attribution reliability and injects qualified signals into the anchors, refining their representations without increasing the token budget.
Extensive evaluations across diverse VLM architectures and benchmarks show that RefineV outperforms state-of-the-art visual token reduction baselines in overall performance. On LLaVA-NeXT-7B, RefineV retains \textbf{96.4\%} of full-token performance while pruning \textbf{94.4\%} of visual tokens. Across the evaluated settings, it achieves up to a \textbf{2.30$\times$} end-to-end inference speedup over unpruned models.

\end{abstract}

\section{Introduction}

Large Vision Language Models (VLMs) \cite{zhu2025internvl3,bai2023qwen} have rapidly evolved from projector-based architectures to high-resolution and unified image--video systems, such as LLaVA-NeXT \cite{liu2024llavanext}, InternVL \cite{chen2024internvl}, and Qwen-VL \cite{wang2024qwen2}. While dense visual tokens improve perception, they also increase attention cost, KV-cache memory, and inference latency \cite{yang2025topv,zhang2024modality}, making visual-token reduction essential for efficient VLM inference.

\begin{figure}[t]
    \centering
    \includegraphics[scale=0.542]{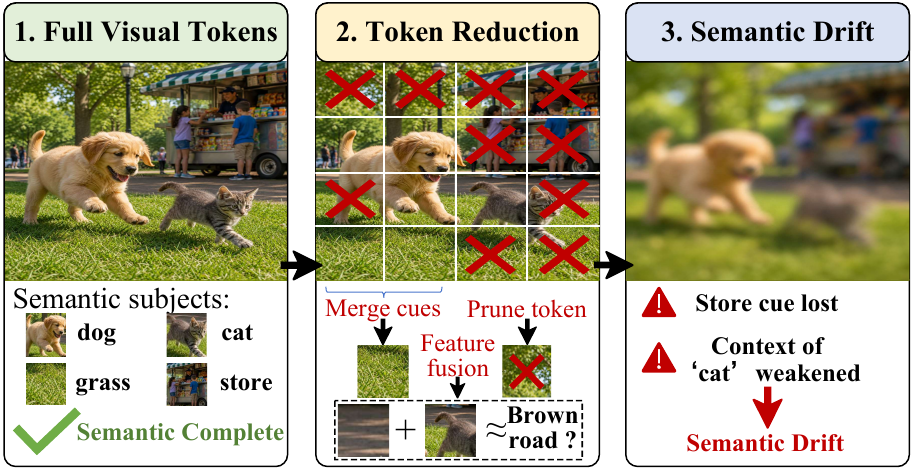} 
    \vspace{0mm}
    \includegraphics[scale=0.349]{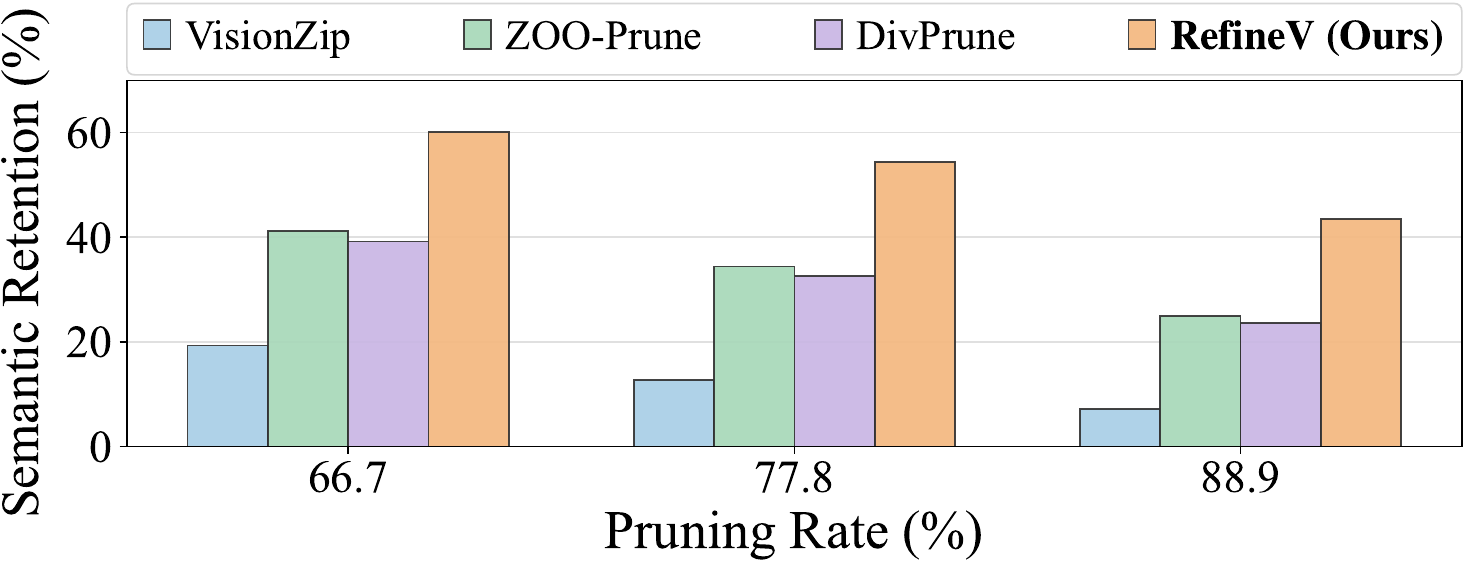} 
    \vspace{-12pt}
    \caption{Illustration and comparison of semantic drift in visual token reduction methods.}
    \vspace{-4pt}
    \label{figure1}
\end{figure}

\begin{figure*}[t]
\centering
\includegraphics[scale=1]{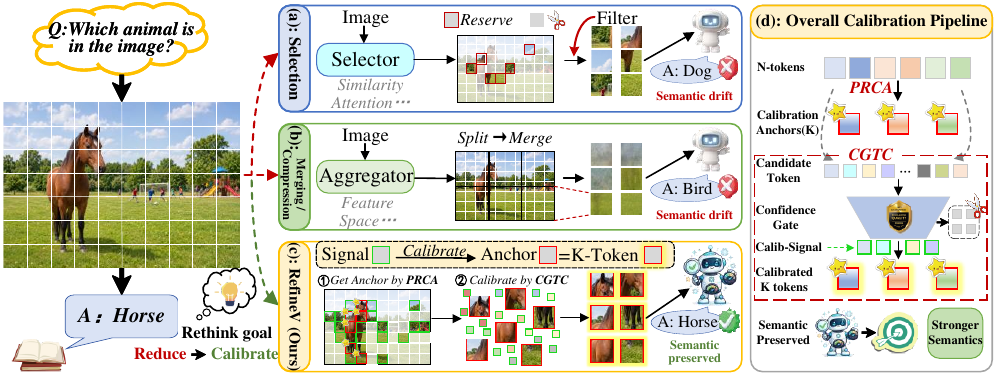}
\vspace{-3pt}
\caption{From token reduction to semantic refinement. (a,b) Selection and merging/compression may discard or dilute critical visual cues. (c) RefineV couples token reduction with semantic refinement before reasoning. (d) Robust anchor selection and confidence-gated residual injection produce $K$ semantically refined visual tokens without increasing the token budget.}
\vspace{-6pt}
\label{figure2}
\end{figure*}

Recent visual token reduction methods follow two routes \cite{lee2026frequency}: pruning tokens according to task- or model-specific importance scores \cite{kim2026zoo,gao2026quietprune,zhang2026d2pruner} and merging or compressing redundant tokens into compact representations \cite{han2026filter,yang2025visionzip,zhang2026hybrid}. Although effective for VLM efficiency, both routes remain token-centric and overlook whether the reduced representation preserves the semantics of the full visual input. We define this failure as \textbf{semantic drift}. As shown in the upper panel of Figure \ref{figure1} and Figure \ref{figure2} (a) and (b), token selection may discard essential cues, whereas compression may entangle redundant information within retained tokens, weakening their semantic specificity.

Prior studies show that pruning loses part of the original visual information, while visual tokens remain critical to language reasoning \cite{zhang2025sparsevlm,yang2025visionzip}. Token reduction should therefore be accompanied by semantic refinement before reasoning, so that compact representations preserve task-relevant visual evidence. We introduce \textbf{RefineV}, a training-free framework that couples visual token reduction with semantic refinement through calibration of the retained representations. The lower panel of Figure~\ref{figure1} shows higher semantic retention for RefineV than for selection and compression baselines. This representation-level diagnostic estimates the coverage of question-relevant source information, rather than directly measuring answer correctness. Appendix A details its calculation and interpretation.

As shown in Figure~\ref{figure2}(c,d), RefineV comprises two complementary modules. 
First, \textbf{Perturbation-Robust Calibration Anchoring} applies multi-directional perturbations before the projector and measures the resulting responses to estimate each token's model-side influence, following sensitivity-based studies \cite{kim2026zoo,sogi2024object}. A directional risk correction favors anchors with consistently strong responses across sampled perturbation directions. A diversity constraint further prevents the selected anchors from concentrating on a single salient region, providing stable reference tokens for subsequent semantic refinement.

Second, \textbf{Confidence-Gated Token Calibration determines} which residual tokens provide reliable complementary information. Its confidence gate combines semantic-spatial affinity, best-anchor matching quality, and assignment concentration to evaluate attribution reliability before information transfer. Qualified signals are then softly assigned and injected into the selected anchors, refining their representations without increasing the token budget. Together, they reduce sequence length while preserving useful evidence from unselected tokens.

Our main contributions are as follows: 

\begin{itemize}
    \item We propose \textbf{RefineV}, a training-free framework that couples visual token reduction with semantic refinement. Building on sensitivity-based selection and information recycling, it addresses anchor-score stability and reliable residual-signal admission.
    \item We propose \textbf{Perturbation-Robust Calibration Anchoring}, which identifies stable, diverse calibration anchors by measuring robust model-side influence.
    \item We design \textbf{Confidence-Gated Token Calibration}, which extracts reliable calibration signals from unselected tokens and injects them into anchors to enrich and calibrate the retained anchor representations.
    \item Extensive experiments show superior overall performance across diverse VLM architectures and token budgets. On LLaVA-NeXT-7B, RefineV retains \textbf{96.4\%} of full-token performance while pruning \textbf{94.4\%} of visual tokens. Across the evaluated settings, it achieves up to \textbf{2.30$\times$} end-to-end speedup.
\end{itemize}

\section{Related Work}

\subsection{Vision-Language Models}

Large Vision-Language Models (VLMs) \cite{zhang2026cofactgvr,ji2025visual,dai2023instructblip,wang2026primt} have achieved remarkable progress in multimodal perception and reasoning by aligning visual encoders with large language models. Early representative systems, such as LLaVA \cite{liu2023visual}, connect a pre-trained vision encoder to an instruction-tuned language model through a lightweight projector, establishing a simple yet effective paradigm for visual instruction tuning. This architecture has been further extended by recent VLM families, including LLaVA-NeXT \cite{liu2024llavanext}, InternVL \cite{chen2024internvl}, and Qwen-VL \cite{wang2024qwen2}, which improve visual understanding through stronger vision backbones, higher-resolution inputs, and more powerful multimodal alignment \cite{zhang2025iebaker}. Despite their effectiveness, these models usually represent each image with hundreds or even thousands of visual tokens \cite{zhang2024user}, leading to substantial computation and memory costs during inference.

\subsection{Visual Token Reduction}

\begin{figure*}[t]
\centering
\includegraphics[scale=0.9]{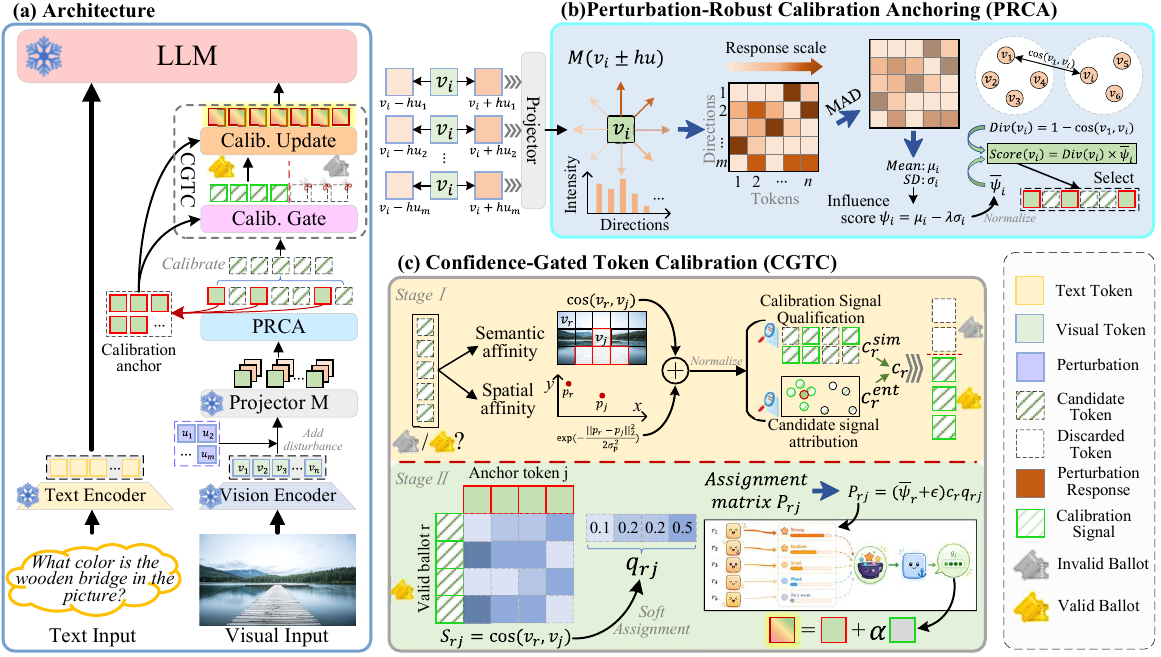}
\caption{\textbf{Overview of RefineV.} (a) PRCA and CGTC couple visual token reduction with semantic refinement. (b) PRCA selects diverse anchors with stable multi-directional projector responses. (c) CGTC first admits reliable residual signals through confidence gating, then softly assigns and injects them into the anchors, yielding $K$ refined visual tokens.}
\label{figure3}
\end{figure*}

Visual token reduction has become an important direction for efficient VLM inference, aiming to reduce the number of visual tokens fed into the language model. Existing methods can be broadly categorized into two paradigms: token selection and token merging/compression \cite{lee2026frequency}. The first paradigm estimates token importance and selects informative tokens based on attention scores, token similarity, diversity, region relevance, or adaptive strategies \cite{kim2026zoo,wang2026posprune}. FastV \cite{chen2024image} and FasterVLM \cite{zhang2024cls} exploit attention statistics for token pruning, SparseVLM \cite{zhang2025sparsevlm} introduces text-guided attention for sparsification, and MADTP \cite{cao2024madtp} leverages multimodal alignment-guided dynamic pruning. HiRED \cite{arif2025hired}, PyramidDrop \cite{xing2024pyramiddrop}, and DivPrune \cite{alvar2025divprune} further explore hierarchical redundancy, diversity preservation, and dynamic selection. The second paradigm reduces tokens through merging or compression by aggregating similar tokens or transforming local visual features into compact representations \cite{wu2026metom,tong2025flashsloth,fu2025framefusion}. Representative methods include LLaVA-PruMerge \cite{shang2025llava}, VisionZip \cite{yang2025visionzip}, TokenPacker \cite{li2025tokenpacker}, and LLaVA-Mini \cite{zhang2025llava}, which explore adaptive merging, dominant token preservation, visual projection and modality fusion.

Despite their effectiveness, existing methods primarily optimize compactness through token selection, merging, or compression \cite{han2026filter}. Critical visual cues may still be discarded, diluted, or incorrectly fused, causing semantic drift. RefineV addresses this issue by coupling token reduction with semantic refinement before reasoning. Rather than treating the reduced sequence as a final representation, it constructs perturbation-robust anchors and calibrates them with selectively admitted residual signals. Its confidence gate screens unreliable sources before soft assignment, while residual injection refines anchor semantics without increasing the token budget. This formulation preserves the efficiency objective of token reduction while explicitly addressing the semantic fidelity of the compact representation.

\section{Methodology}

In this section, we present RefineV, a training-free framework that couples fixed-budget visual token reduction with semantic refinement. As illustrated in Figure~\ref{figure3}, PRCA selects anchors with stable model-side influence, while CGTC screens residual tokens according to attribution reliability and injects qualified signals to refine those anchors. ``Robust'' refers specifically to score stability across sampled perturbation directions, rather than general adversarial robustness.

\subsection{Preliminary and Problem Formulation}
Given an input image, the vision encoder produces a sequence of visual tokens $V=[v_1,\ldots,v_N]^\top \in {R}^{N\times d_v}$, where $N$ is the number of visual tokens and $d_v$ is the token dimension. For a fixed retention ratio $\rho$ , the target budget is $K=\max(1,\lfloor \rho N \rfloor)$. Instead of directly selecting or compressing the original sequence into $K$ tokens, our goal is to generate a calibrated visual representation $\hat{V}_S=\{\hat{v}_j\mid j\in S\}$, where $S\subset\{1,\ldots,N\}$ and cardinality $|S| = K$. The calibrated tokens preserve usable semantics from the full visual input while satisfying the same ratio $\rho$.

Prior sensitivity-based pruning studies \cite{kim2026zoo,sogi2024object} suggest that perturbing visual tokens before the multimodal projector reflects their influence on model-side visual mappings. Following this observation, we utilize the projector $M(\cdot)$ as a lightweight response function. RefineV needs to address two key questions: which visual tokens serve as stable semantic reference points, and how unselected visual tokens should contribute to calibration. These questions are addressed by Perturbation-Robust Calibration Anchoring (PRCA) and Confidence-Gated Token Calibration (CGTC), respectively.

\subsection{Perturbation-Robust Calibration Anchoring}
As shown in Figure \ref{figure3}(b), for visual tokens $V$, we sample the same $m$ unit perturbation directions $\{u_x\}_{x=1}^{m}$, obtained by sampling $g_x\sim\mathcal{N}(0,I_{d_v})$ and normalizing it as $u_x=g_x/\lVert g_x\rVert_2$, and apply symmetric perturbations before the projector with step size $h$. The response of token $v_i$ along direction $x$ is computed as:
\begin{equation}
C_{x,i}=\frac{\|M(v_i+h u_{x})-M(v_i-hu_{x})\|_2}{2h},
\label{eq1}
\end{equation}
where $\|\cdot\|_2$ denotes the $\ell_2$-norm.

This response measures how strongly the model-side visual representation changes when a token is locally perturbed. However, different perturbation directions may have different response scales. Directly averaging raw responses may overemphasize a few high-scale directions. Therefore, we perform direction-wise robust normalization with the median and median absolute deviation:
\begin{equation}
Z_{x,i}=\frac{C_{x,i}-\mathrm{med}_{j}(C_{x,j})}{\mathrm{med}_{j}|C_{x,j}-\mathrm{med}_{q}(C_{x,q})|+\epsilon},
\label{eq2}
\end{equation}
where $\epsilon$ is a small positive constant, $\mathrm{med}_{j}(\cdot)$ denotes the median taken over index $j$. 

For each token, we compute the mean response $\mu_i=\frac{1}{m}\sum_x Z_{x,i}$ and the directional standard deviation $\sigma_i=\left(\frac{1}{m}\sum_x (Z_{x,i}-\mu_i)^2\right)^{1/2}$. If token with strong responses in only a few perturbation directions may be an unstable anchor. We therefore define a perturbation-robust influence score as $\psi_i=\mu_i-\lambda\sigma_i$, where $\lambda$ controls the penalty for directional risk. Larger $\psi_i$ indicates that the token consistently influences the model-side mapping across perturbation directions. We then normalize $\psi_i$ to $\bar{\psi}_i\in[0,1]$ before anchor selection.

Selecting anchors solely by $\bar{\psi}_i$ may lead to many anchors concentrated in the same salient object or local region. To avoid this, PRCA further incorporates diversity. We define the visual distance between two tokens as $D(v_i,v_j)=1-\cos(v_i,v_j)$. The anchor set is initialized with the token of the largest $\bar{\psi}_i$. At step $t$, for each unselected token $v_i$, we compute its marginal diversity to the current anchor set as $\delta_i^{(t)}=\min_{j\in S_t}D(v_i,v_j)$, and select the next anchor by
\begin{equation}
i_{t+1}=\arg\max_{i\notin S_t}\bar{\psi}_i\delta_i^{(t)}, \quad S_{t+1}=S_t\cup\{i_{t+1}\}.
\label{eq3}
\end{equation}

This process is repeated until $|S| = K$. The selected set $S$ forms the calibration anchors, and the remaining tokens $R = \left\{ i \in \{1,\ldots,N\} \mid i \notin S \right\}$ are treated as candidate tokens for subsequent calibration.

\subsection{Confidence-Gated Calibration Signals}
CGTC determines which unselected tokens provide reliable calibration signals. Not every residual token should be fused into anchors: some may correspond to noisy background, or regions that cannot be confidently assigned. Therefore, the first stage of CGTC qualifies candidate calibration signals through a confidence gate, as shown in Figure \ref{figure3}(c).

For each residual token $\{v_r | r\in R\}$ and anchor token $\{v_j | j\in S\}$, we compute a semantic-spatial affinity:
\begin{equation}
A_{rj}=\cos(v_r,v_j)+\eta\exp\left(-\frac{\|p_r-p_j\|_2^2}{2\sigma_p^2}\right),
\label{eq4}
\end{equation}
where $p_r$ and $p_j$ are the 2D patch coordinates of tokens $v_r$ and $v_j$, $\eta$ controls the spatial term, $\sigma_p$ controls the spatial decay. The cosine term measures visual semantic similarity, while the spatial term encourages locally coherent assignments.

For each residual token, we take its top-$k_f$ ($k_f = \min\left(K,\max\left(2,\left\lceil 0.1K \right\rceil\right)\right)$) anchors according to $A_{rj}$ and compute a normalized assignment distribution $\pi_r$ with temperature $\tau_g$. Then evaluate two confidence scores. The first is the best-anchor matching confidence, defined by $c_r^{\mathrm{sim}}=\mathrm{sigmoid}\left(\frac{A_r^{(1)}-\theta_s}{\tau_c}\right)$, where $A_r^{(1)}$ is the maximum affinity, $\theta_s$ is affinity threshold. The second is the assignment concentration confidence, which penalizes ambiguous assignments:
\begin{equation}
c_r^{\mathrm{ent}}=1+\frac{\sum_{\ell=1}^{k_f}\pi_{r\ell}\log \pi_{r\ell}}{\log k_f}.
\label{eq5}
\end{equation}

If a residual token is evenly distributed over multiple anchors, its entropy is high and $c_r^{\mathrm{ent}}$ becomes small. We combine these two views as:
\begin{equation}
c_r=c_r^{\mathrm{sim}}\frac{1+c_r^{\mathrm{ent}}}{2}, \quad R^+=\{r\in R\mid c_r\geq\theta_c\},
\label{eq6}
\end{equation}
where $\theta_c$ is confidence threshold. Only tokens in $R^+$ are regarded as valid calibration signals. Their source weights are defined as $w_r=(\bar{\psi}_r+\epsilon)c_r$, where $\bar{\psi}_r$ measures model-side influence and $c_r$ measures attribution reliability.

\subsection{Anchor Calibration with Signals}
The second stage of CGTC injects the qualified calibration signals into their matched anchors while keeping the final token budget unchanged. For each valid residual token $\{v_r|r \in R^{+}\}$, we compute its soft assignment to all calibration anchors by:
\begin{equation}
q_{rj}=\frac{\exp(\cos(v_r,v_j)/\tau_s)}{\sum_{l\in S}\exp(\cos(v_r,v_l)/\tau_s)}, \quad j\in S.
\label{eq7}
\end{equation}

The assignment matrix is then weighted by the source reliability, $P_{rj}=w_r q_{rj}$. For each anchor token $v_j$, the aggregated calibration signal is computed as a normalized weighted average:
\begin{equation}
g_j=\frac{\sum_{r\in R^+}P_{rj}v_r}{\sum_{r\in R^+}P_{rj}+\epsilon}.
\label{eq8}
\end{equation}

\begin{table*}[t]
\centering
\fontsize{9}{8}\selectfont
\setlength{\tabcolsep}{4.9pt}
\renewcommand{\arraystretch}{0.95}

\begin{tabular}{l|c|ccccccccc|c}
\specialrule{0.8pt}{0pt}{0pt}
\textbf{Method} 
&\textbf{TFLOPs}
& \makecell{\textbf{GQA}}
& \makecell{\textbf{MMB}}
& \makecell{\textbf{MME}}
& \makecell{\textbf{POPE}}
& \makecell{\textbf{SQA}}
& \makecell{\textbf{VQA}$^{\mathbf{v2}}$}
& \makecell{\textbf{VQA}$^{\mathbf{T}}$}
& \makecell{\textbf{MMMU}}
& \makecell{\textbf{SEED}}
& \makecell{\textbf{Avg.}} \\
\specialrule{0.8pt}{0pt}{0pt}

\rowcolor{gray!20}
\multicolumn{12}{c}{\textit{Total 576 Tokens}} \\
\hline
LLaVA-1.5-7B 
&8.54 & 61.90 & 64.70 & 1862.00 & 85.90 & 69.50 & 78.50 & 58.20 & 36.30 & 58.60 & 100\% \\
\hline

\rowcolor{gray!20}
\multicolumn{12}{c}{\textit{Retain 192 Tokens} \textcolor{green!70!black}{$\downarrow$ 66.7\%}} \\
\hline
FastV (ECCV 2024) 
& 3.44 & 52.70 & 61.20 & 1612.00 & 64.80 & 67.30 & 67.10 & 52.50 & 34.30 & 57.10 & 89.6\% \\

SparseVLM (ICML 2025) 
& 3.46 & 57.60 & 62.50 & 1721.00 & 83.60 & 69.10 & 75.60 & 56.10 & 33.80 & 55.80 & 95.5\% \\

VisionZip (CVPR 2025) 
& 3.93 & 59.30 & 63.00 & 1782.00 & 85.30 & 68.90 & 76.80 & 57.30 & 36.60 & 56.40 & 97.9\% \\

DivPrune (CVPR 2025) 
& - & 59.97 & 62.54 & 1762.30 & 87.00 & 68.91 & 76.87 & 56.97 & 35.44 & 58.71 & 98.0\% \\

FiCoCO-V (AAAI 2026) 
& 3.20 & 58.51 & 61.74 & 1755.61 & 81.47 & 67.42 & 74.22 & 55.50 & 36.11 & 57.95 & 96.0\% \\

V$^2$Drop (CVPR 2026) 
& 3.42 & 58.60 & 62.37 & 1789.76 & 85.10 & 69.30 & 74.92 & 55.62 & 35.87 & 56.74 & 97.0\% \\

ZOO-Prune (CVPR 2026) 
& 3.98 & 60.03 & 62.57 & 1781.66 & 87.24 & 69.01 & 75.35 & 57.10 & 35.44 & 58.78 & 98.0\% \\

\textbf{\textit{RefineV} (Ours)} 
& 3.47 & \textbf{60.57} & \textbf{63.41} & \textbf{1809.47} & \textbf{88.94} & \textbf{70.24} & \textbf{77.42} & \textbf{58.01} & \textbf{37.33} & \textbf{58.94} 
& \textcolor{red!80!black}{\textbf{99.9\%}} \\
\hline

\rowcolor{gray!20}
\multicolumn{12}{c}{\textit{Retain 128 Tokens} \textcolor{green!70!black}{$\downarrow$ 77.8\%}} \\
\hline
FastV (ECCV 2024) 
& 2.62 & 49.60 & 56.10 & 1490.00 & 59.60 & 61.80 & 61.80 & 50.60 & 34.90 & 55.90 & 84.7\% \\

SparseVLM (ICML 2025) 
& 2.90 & 56.00 & 60.00 & 1696.00 & 80.50 & 67.10 & 73.80 & 54.90 & 33.80 & 53.40 & 93.0\% \\

VisionZip (CVPR 2025) 
& 3.15 & 57.60 & 62.00 & \textbf{1761.70} & 83.20 & 68.90 & 75.60 & 56.80 & \textbf{37.90} & 54.90 & 96.8\% \\

DivPrune (CVPR 2025) 
& - & 59.25 & 62.03 & 1718.22 & 86.72 & 68.96 & 75.96 & 56.06 & 35.56 & 56.98 & 96.9\% \\

FiCoCO-V (AAAI 2026) 
& 2.79 & 57.59 & 60.34 & 1718.06 & 82.24 & 68.73 & 72.97 & 54.10 & 35.44 & 54.98 & 94.5\% \\

V$^2$Drop (CVPR 2026) 
& 2.59 & 56.19 & 59.28 & 1712.59 & 80.90 & 68.83 & 73.11 & 53.80 & 35.12 & 55.82 & 93.9\% \\

ZOO-Prune (CVPR 2026) 
& 3.13 & 59.43 & 61.86 & 1741.23 & 87.13 & 68.86 & 74.55 & 56.79 & 35.67 & 57.39 & 97.2\% \\

\textbf{\textit{RefineV} (Ours)} 
& 3.02 & \textbf{60.15} & \textbf{62.11} & 1731.46 & \textbf{87.79} & \textbf{69.89} & \textbf{76.73} & \textbf{57.96} & 36.22 & \textbf{57.73} 
& \textcolor{red!80!black}{\textbf{98.3\%}} \\
\hline

\rowcolor{gray!20}
\multicolumn{12}{c}{\textit{Retain 64 Tokens} \textcolor{green!70!black}{$\downarrow$ 88.9\%}} \\
\hline
FastV (ECCV 2024) 
& 1.98 & 46.10 & 48.00 & 1256.00 & 48.00 & 51.10 & 55.00 & 47.80 & 34.00 & 51.90 & 75.5\% \\

SparseVLM (ICML 2025) 
& 2.11 & 52.70 & 56.20 & 1505.00 & 75.10 & 62.20 & 68.20 & 51.80 & 32.70 & 51.10 & 87.0\% \\

VisionZip (CVPR 2025) 
& 2.25 & 55.10 & 60.10 & \textbf{1690.00} & 77.00 & 69.00 & 72.40 & 55.50 & 35.20 & 52.20 & 92.8\% \\

DivPrune (CVPR 2025) 
& - & 57.78 & 59.28 & 1674.40 & 85.56 & 68.17 & 74.11 & 54.69 & 35.56 & 55.13 & 94.8\% \\

FiCoCO-V (AAAI 2026) 
& 1.68 & 53.43 & 55.09 & 1653.67 & 75.29 & 67.92 & 71.63 & 51.88 & 34.22 & 53.14 & 90.1\% \\

V$^2$Drop (CVPR 2026) 
& 1.76 & 53.52 & 55.21 & 1641.23 & 75.14 & 68.92 & 71.03 & 51.83 & 34.29 & 53.11 & 90.1\% \\

ZOO-Prune (CVPR 2026) 
& 2.29 & 58.36 & 60.22 & 1626.80 & 85.56 & 68.17 & 72.97 & 55.35 & 35.21 & \textbf{55.84} & 94.8\% \\

\textbf{\textit{RefineV} (Ours)} 
& 2.18 & \textbf{59.12} & \textbf{60.58} & 1664.51 & \textbf{86.98} & \textbf{69.80} & \textbf{75.81} & \textbf{56.32} & \textbf{36.11} & 55.78 
& \textcolor{red!80!black}{\textbf{96.5\%}} \\

\specialrule{0.8pt}{0pt}{0pt}
\end{tabular}

{\fontsize{9}{8}\selectfont
\vspace{-3pt}
\caption{Performance comparison on \textbf{LLaVA-1.5-7B}.}
\vspace{-8pt}
\label{tab:llava15_7b}
}

\end{table*}

\begin{table}[t]
\centering
\fontsize{9}{9}\selectfont
\setlength{\tabcolsep}{4.9pt}
\renewcommand{\arraystretch}{1.08}

\begin{tabular}{l|cccc|c}
\specialrule{0.8pt}{0pt}{0pt}
\textbf{Method} 
& GQA & MMB & MME & POPE & Avg.-4 \\
\specialrule{0.8pt}{0pt}{0pt}

\rowcolor{gray!20}
\multicolumn{6}{c}{\textit{Full Tokens}} \\
\hline
Qwen2.5-VL-7B 
& 60.84 & 82.82 & 2310 & 86.30 & 100\% \\
\hline

\rowcolor{gray!20}
\multicolumn{6}{c}{\textit{Retain 20\% Tokens} \textcolor{green!70!black}{$\downarrow$ 80.0\%}} \\
\hline

FiCoCO-V 
& 55.28 & 76.47 & 2120 & 81.29 & 92.3\% \\

ZOO-Prune 
& 55.92 & 77.41 & 2139 & 81.71 & 93.2\% \\

\textbf{\textit{RefineV} (Ours)} 
& \textbf{58.11} & \textbf{77.91} & \textbf{2274} & \textbf{83.94} & \textcolor{red!80!black}{\textbf{96.3\%}} \\
\hline

\rowcolor{gray!20}
\multicolumn{6}{c}{\textit{Retain 10\% Tokens} \textcolor{green!70!black}{$\downarrow$ 90.0\%}} \\
\hline
FiCoCO-V 
& 52.79 & 71.95 & 1967 & 77.61 & 87.2\% \\

ZOO-Prune 
& 53.28 & 72.51 & 1962 & 78.46 & 87.7\% \\

\textbf{\textit{RefineV} (Ours)} 
& \textbf{54.45} & \textbf{73.37} & \textbf{2017} & \textbf{79.23} & \textcolor{red!80!black}{\textbf{89.3\%}} \\
\hline

\specialrule{0.8pt}{0pt}{0pt}
\end{tabular}

{\fontsize{9}{8}\selectfont
\vspace{-3pt}
\caption{Performance comparison on \textbf{Qwen2.5-VL-7B}.}
\vspace{-8pt}
\label{tab:qwen25_7b}
}
\vspace{-1em}
\end{table}

Finally, the anchor representation is updated by $v'_j=v_j+\alpha g_j$, where $\alpha$ controls the calibration strength. To reduce distribution drift at the language-model interface, we preserve the original anchor norm:
\begin{equation}
\hat{v}_j=\frac{\|v_j\|_2}{\|v'_j\|_2+\epsilon}v'_j.
\label{eq9}
\end{equation}

The resulting $\hat{V}_S=\{\hat{v}_j\mid j\in S\}$ contains exactly $K$ calibrated visual tokens, which are combined with text tokens for subsequent LLM reasoning through the original model interface. RefineV thus retains a fixed visual-token budget while using reliable evidence from unselected tokens to refine the compact visual representations.

\section{Experiments}

\begin{table*}[t]
\centering
\fontsize{9}{8}\selectfont
\setlength{\tabcolsep}{7pt}
\renewcommand{\arraystretch}{0.95}

\begin{tabular}{l|ccccccccc|c}
\specialrule{0.8pt}{0pt}{0pt}
\textbf{Method} 
& \makecell{\textbf{GQA}}
& \makecell{\textbf{MMB}}
& \makecell{\textbf{MME}}
& \makecell{\textbf{POPE}}
& \makecell{\textbf{SQA}}
& \makecell{\textbf{VQA}$^{\mathbf{v2}}$}
& \makecell{\textbf{VQA}$^{\mathbf{T}}$}
& \makecell{\textbf{MMMU}}
& \makecell{\textbf{SEED}}
& \makecell{\textbf{Avg.}} \\
\specialrule{0.8pt}{0pt}{0pt}

\rowcolor{gray!20}
\multicolumn{11}{c}{\textit{Total 2880 Tokens}} \\
\hline
LLaVA-NeXT-7B 
& 64.20 & 67.90 & 1842.00 & 86.40 & 70.20 & 80.10 & 61.30 & 35.10 & 70.20 & 100\% \\
\hline

\rowcolor{gray!20}
\multicolumn{11}{c}{\textit{Retain 640 Tokens} \textcolor{green!70!black}{$\downarrow$ 77.8\%}} \\
\hline

VisionZip (CVPR 2025) 
& 61.30 & \textbf{66.30} & 1787.00 & 86.30 & 68.10 & \textbf{79.10} & \textbf{60.20} & 34.70 & 66.70 & 97.5\% \\

FiCoCO-V (AAAI 2026) 
& 60.40 & 64.18 & 1780.22 & 86.04 & 67.94 & 75.77 & 54.89 & 35.91 & 65.77 & 95.8\% \\

ZOO-Prune (CVPR 2026) 
& 62.19 & 65.10 & 1802.47 & 86.75 & 68.02 & 78.13 & 57.98 & 36.82 & 67.59 & 97.9\% \\

\textbf{\textit{RefineV }(Ours)} 
& \textbf{63.87} & 65.98 & \textbf{1831.14} & \textbf{88.19} & \textbf{69.41} & 78.02 & 58.39 & \textbf{37.89} & \textbf{67.92} & \textcolor{red!80!black}{\textbf{99.4\%}} \\
\hline

\rowcolor{gray!20}
\multicolumn{11}{c}{\textit{Retain 320 Tokens} \textcolor{green!70!black}{$\downarrow$ 88.9\%}} \\
\hline

VisionZip (CVPR 2025) 
& 59.30 & 63.10 & 1702.00 & 82.10 & 67.30 & 76.20 & \textbf{58.90} & 35.30 & 63.40 & 94.5\% \\

FiCoCO-V (AAAI 2026) 
& 59.26 & 63.30 & 1721.74 & 85.18 & 67.51 & 74.08 & 54.67 & 34.33 & 64.51 & 93.9\% \\

ZOO-Prune (CVPR 2026) 
& 60.83 & 64.60 & 1787.68 & 85.47 & 67.08 & 76.04 & 57.26 & 36.78 & 66.18 & 96.5\% \\

\textbf{\textit{RefineV }(Ours)} 
& \textbf{61.98} & \textbf{64.92} & \textbf{1791.90} & \textbf{87.58} & \textbf{68.86} & \textbf{77.76} & 57.67 & \textbf{37.33} & \textbf{66.28} 
& \textcolor{red!80!black}{\textbf{97.9\%}} \\
\hline

\rowcolor{gray!20}
\multicolumn{11}{c}{\textit{Retain 160 Tokens} \textcolor{green!70!black}{$\downarrow$ 94.4\%}} \\
\hline

VisionZip (CVPR 2025) 
& 55.50 & 60.10 & 1630.00 & 74.80 & 68.30 & 71.40 & \textbf{56.20} & 36.10 & 58.30 & 90.4\% \\

FiCoCO-V (AAAI 2026) 
& 58.11 & 60.32 & 1657.81 & 84.27 & 68.01 & 73.22 & 55.10 & 33.98 & 63.06 & 92.4\% \\

ZOO-Prune (CVPR 2026) 
& 59.48 & 64.18 & 1706.08 & 83.05 & 67.72 & 74.07 & 55.42 & \textbf{36.67} & 64.04 & 94.5\% \\

\textbf{\textit{RefineV }(Ours)} 
& \textbf{59.97} & \textbf{64.47} & \textbf{1706.96} & \textbf{86.77} & \textbf{69.70} & \textbf{75.81} & 55.77 & 36.56 & \textbf{67.87}
& \textcolor{red!80!black}{\textbf{96.4\%}} \\

\specialrule{0.8pt}{0pt}{0pt}
\end{tabular}

{\fontsize{9}{8}\selectfont
\vspace{-4pt}
\caption{Performance comparison on \textbf{LLaVA-NeXT-7B}. Comprehensive comparison is provided in the Supplementary Sec. C.}
\vspace{-2pt}
\label{tab:llava16_7b}
}
\vspace{-0.5em}
\end{table*}

We evaluate RefineV on LLaVA-1.5-7B \cite{liu2024improved}, LLaVA-NeXT-7B \cite{liu2024llavanext} across nine benchmarks, and in eight benchmarks for Qwen2.5-VL-7B \cite{bai2025qwen25vltechnicalreport}. LLaVA-1.5-7B encodes images into 576 tokens. LLaVA-NeXT represents each image with 2,880 tokens, and Qwen2.5-VL adopts a dynamic-resolution ViT encoder. We assess RefineV and competing methods by LMMs-Eval \cite{zhang2025lmms}. All experiments are conducted on a single NVIDIA GeForce RTX 5090 D GPU with $h=5\times10^{-4}$, $m=64$, $\lambda=0.5$, $\eta=0.45$, $\theta_s=0.4$, $\theta_c=0.9$, and $\alpha=0.15$. The compared methods span two categories of token reduction. Token selection methods: FastV \cite{chen2024image}, SparseVLM \cite{zhang2025sparsevlm}, DivPrune \cite{alvar2025divprune}, V$^2$Drop \cite{chen2026variation} and ZOO-Prune \cite{kim2026zoo}; and token compression methods: VisionZip \cite{yang2025visionzip} and FiCoCo \cite{han2026filter}. Further details are provided in Supplementary Sec. C. 

\begin{figure}[t]
\centering
\includegraphics[scale=0.775]{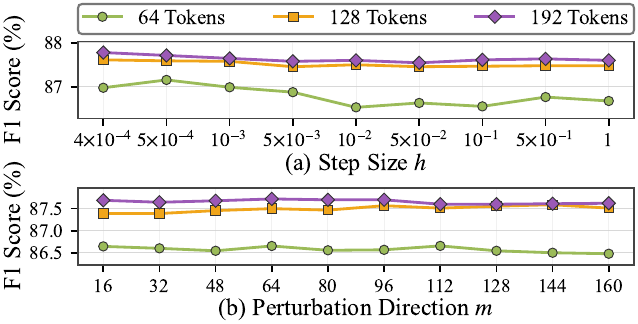}
\vspace{-12pt}
\caption{Ablation with hyperparameters ``$h$'' and ``$m$''. }
\vspace{-8pt}
\label{figure6}
\end{figure}

\subsection{Comparison on Diverse Tasks}

\begin{table*}[t]
\centering
\fontsize{9}{8}\selectfont
\setlength{\tabcolsep}{8.3pt}
\renewcommand{\arraystretch}{0.95}

\begin{tabular}{cc|ccccccccc|c}
\specialrule{0.8pt}{0pt}{0pt}
\textbf{PRCA} & \textbf{CGTC}
& \makecell{\textbf{GQA}}
& \makecell{\textbf{MMB}}
& \makecell{\textbf{MME}}
& \makecell{\textbf{POPE}}
& \makecell{\textbf{SQA}}
& \makecell{\textbf{VQA}$^{\mathbf{v2}}$}
& \makecell{\textbf{VQA}$^{\mathbf{T}}$}
& \makecell{\textbf{MMMU}}
& \makecell{\textbf{SEED}}
& \makecell{\textbf{Avg.}} \\
\specialrule{0.8pt}{0pt}{0pt}

\rowcolor{gray!20}
\multicolumn{12}{c}{\textit{Retain 192 Tokens} \textcolor{green!70!black}{$\downarrow$ 66.7\%}} \\
\hline
 - & -
& 60.03 & 62.57 & 1781.66 & 87.24 & 69.01 & 75.35 & 57.10 & 35.44 & 58.78 & 98.0\% \\

$\checkmark$ & -
& 60.11 & 63.01 & 1784.31 & 87.78 & 69.80 & 77.01 & 57.19 & 36.56 & 58.89 & 99.0\% \\

- & $\checkmark$
& 60.52 & 63.23 & 1801.97 & 88.72 & 68.77 & 75.46 & 57.49 & 35.98 & 58.87 & 98.8\% \\

$\checkmark$ & $\checkmark$
& \textbf{60.57} & \textbf{63.41} & \textbf{1809.47} & \textbf{88.94} & \textbf{70.24} & \textbf{77.42} & \textbf{58.01} & \textbf{37.33} & \textbf{58.94} 
& \textcolor{red!80!black}{\textbf{99.9\%}} \\
\hline

\rowcolor{gray!20}
\multicolumn{12}{c}{\textit{Retain 128 Tokens} \textcolor{green!70!black}{$\downarrow$ 77.8\%}} \\
\hline
 - & -
& 59.43 & 61.86 & 1741.23 & 87.13 & 68.86 & 74.55 & 56.79 & 35.67 & 57.39 & 97.2\% \\

$\checkmark$ & -
& 59.72 & 61.36 & 1733.30 & 87.50 & 69.42 & \textbf{76.73} & 56.62 & 36.11 & 57.46 & 97.6\% \\

- & $\checkmark$
& 59.87 & 62.00 & \textbf{1747.59} & 87.58 & 68.52 & 74.76 & 56.96 & 35.89 & 57.55 & 97.5\% \\

$\checkmark$ & $\checkmark$
& \textbf{60.15} & \textbf{62.11} & 1731.46 & \textbf{87.79} & \textbf{69.89} & \textbf{76.73} & \textbf{57.96} & \textbf{36.22} & \textbf{57.73} 
& \textcolor{red!80!black}{\textbf{98.3\%}} \\
\hline

\rowcolor{gray!20}
\multicolumn{12}{c}{\textit{Retain 64 Tokens} \textcolor{green!70!black}{$\downarrow$88.9\%}} \\
\hline
 - & -
& 58.36 & 60.22 & 1626.80 & 85.56 & 68.17 & 72.97 & 55.35 & 35.21 & \textbf{55.84} & 94.8\% \\

$\checkmark$ & -
& 58.61 & 60.23 & 1631.64 & 86.28 & 68.71 & 75.61 & 56.26 & 35.44 & 55.46 & 95.6\% \\

- & $\checkmark$
& 58.95 & 60.57 & 1648.61 & 86.53 & 68.52 & 73.25 & 55.71 & 35.56 & 55.71 & 95.4\% \\

$\checkmark$ & $\checkmark$
& \textbf{59.12} & \textbf{60.58} & \textbf{1664.51} & \textbf{86.98} & \textbf{69.80} & \textbf{75.81} & \textbf{56.32} & \textbf{36.11} & 55.78
& \textcolor{red!80!black}{\textbf{96.5\%}} \\

\specialrule{0.8pt}{0pt}{0pt}
\end{tabular}

{\fontsize{9}{8}\selectfont
\vspace{-3pt}
\caption{Module ablation on LLaVA-1.5-7B. Further details are provided in Supplementary Sec. D.}
\vspace{-4pt}
\label{tab:module_ablation}
}
\vspace{-1.5em}
\end{table*}

\paragraph{Results on LLaVA-1.5-7B.}
As shown in Table~\ref{tab:llava15_7b}, our RefineV consistently outperforms existing selection- and compression-based pruning methods. With only 192 tokens, it retains 99.9\% of the unpruned performance. And it surpasses ZOO-Prune \cite{kim2026zoo} by 1.9, 1.1, and 1.7 percentage points at three pruning rates, while maintaining low TFLOPs. Further analysis and TFLOPs details of calculation are provided in Supplementary Sec. B and C.
\paragraph{Results on Qwen2.5-VL-7B.}
To verify the generalization capability of RefineV, we further evaluate it on the new Qwen-series model, Qwen2.5-VL-7B, as shown in Table~\ref{tab:qwen25_7b}. With only 20\% and 10\% of the tokens retained, it achieves 96.3\% and 89.3\% performance retention, respectively, consistently outperforming Comparison methods \cite{kim2026zoo,han2026filter}. This further shows superior performance over conventional token reduction methods with long, variable-length token sequences. We present partial results here; the complete results are available in Supplementary Sec. C.
\paragraph{Results on LLaVA-NeXT-7B.}
As shown in Table~\ref{tab:llava16_7b}, we further evaluate RefineV on LLaVA-NeXT-7B for high-resolution images. Even at 94.4\% pruning rate, it preserves 96.4\% performance and surpasses the second-best method by 1.5, 1.4, and 1.9 percentage points across three pruning rates, demonstrating the superiority of token calibration over conventional token reduction methods.

\subsection{Ablation Studies and Further Analyses}
\paragraph{Ablation Studies.}

We conduct ablation studies on the two modules of RefineV, PRCA and CGTC, across nine benchmarks, as shown in Table~\ref{tab:module_ablation}. When PRCA is removed, we adopt the existing token selection method ZOO-Prune \cite{kim2026zoo} as the baseline. Table \ref{tab:module_ablation} clearly shows that adding either PRCA or CGTC consistently improves performance, while combining both leads to more substantial performance gains over the baseline. These findings show that the two modules are both effective and complementary for visual token calibration. Their combination achieves the best results. Further analysis is provided in Supplementary Sec. D.

\paragraph{Impact of $h$ and $m$.}
We analyze the perturbation step size $h$ and number of perturbation directions $m$ on POPE with LLaVA-1.5-7B. Across three cases, $h=5\times10^{-4}$ and $m=64$ yield high F1 scores. Since lower values of $m$ reduce computation, we adopt these settings for accuracy--efficiency balance, as shown in Figure \ref{figure6}. 
\begin{figure*}[t]
\centering
\includegraphics[scale=0.733]{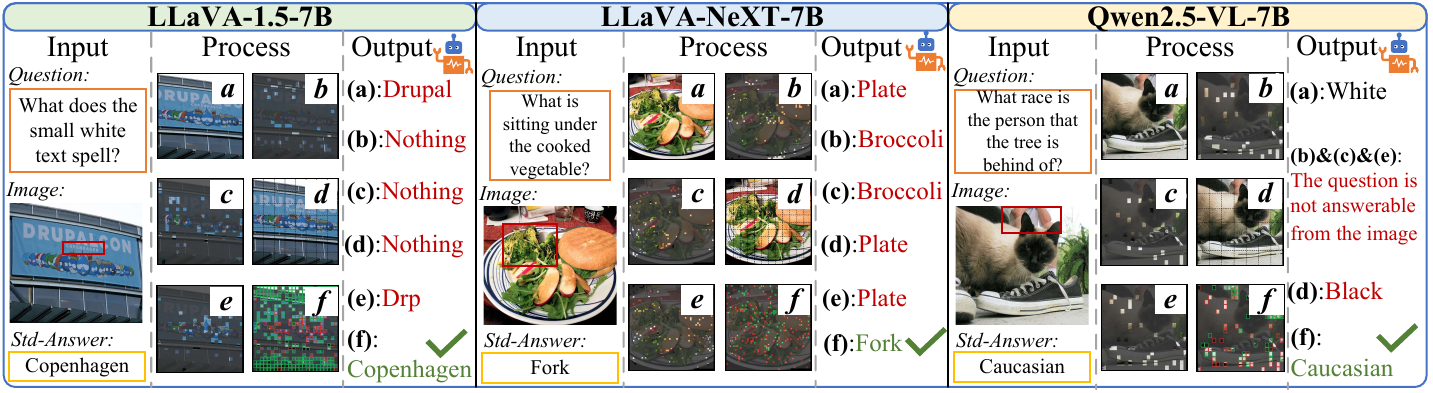}
\vspace{-13pt}
\caption{Qualitative visualizations. (a) Baseline; (b) sensitivity-based token selection; (c) similarity-based token selection; (d) token compression via feature merging; (e) PRCA-only retaining selected Calib-Anchors; (f) \textbf{RefineV (Ours)}. ``Process'' shows image processing; red and green boxes in (f) mark Calib-Anchors and calibration signals. See Supplementary Sec. E for details. }
\vspace{-4pt}
\label{figure7}
\end{figure*}

\begin{figure}[t]
\centering
\includegraphics[scale=0.473]{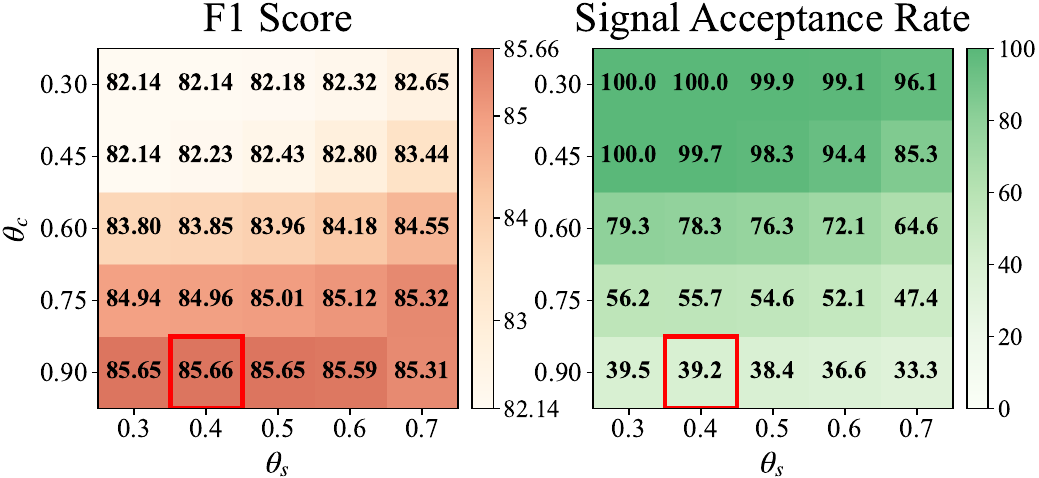}
\vspace{-4pt}
\caption{Heatmap-based ablation study of $\theta_s$ and $\theta_c$ on POPE with LLaVA-1.5-7B at a pruning rate of 94.4\%. }
\vspace{-2pt}
\label{figure8}
\end{figure}

\paragraph{Impact of $\theta_s$ and $\theta_c$.}
Figure~\ref{figure8} examines affinity and confidence thresholds for residual-signal admission on POPE with LLaVA-1.5-7B. At $\theta_s=0.4$, raising $\theta_c$ from 0.30 to 0.90 improves F1 from 82.14 to 85.66 while reducing acceptance from 100.0\% to 39.2\%. Intermediate thresholds of 0.60 and 0.75 yield F1 scores of 83.85 and 84.96 with acceptance rates of 78.3\% and 55.7\%, respectively, supporting selective residual filtering. At $\theta_c=0.90$, $\theta_s\in\{0.3,0.4\}$ yields similar F1 scores (85.65--85.66), whereas $\theta_s=0.7$ reduces acceptance to 33.3\% and F1 to 85.31, suggesting that excessive filtering removes useful complementary evidence. These results favor selective admission over either extreme. We adopt $(\theta_s,\theta_c)=(0.4,0.9)$, achieving the highest displayed F1 while retaining a substantial subset of residual signals. Appendix.D provides further analysis.

\begin{figure}[t]
\centering
\includegraphics[scale=0.5]{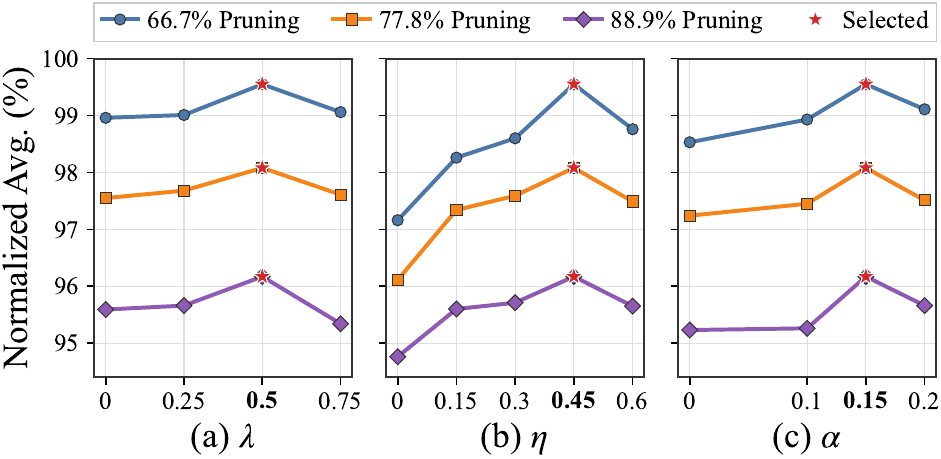} 
\vspace{-6pt}
\caption{Ablation study of $\lambda$, $\eta$ and $\alpha$ on the normalized average over six
benchmarks with LLaVA-1.5-7B. }
\vspace{-7pt}
\label{figure9}
\end{figure}

\paragraph{Impact of $\lambda$, $\eta$ and $\alpha$.}
As shown in Figure \ref{figure9}, we conduct ablation studies on LLaVA-1.5-7B under three pruning ratios across six benchmarks: $GQA$, $MMB$, $MME$, $POPE$, $SQA$, and $VQA^T$. We evaluate the robustness regularization strength $\lambda$ used in Calib-Anchor computation, the spatial weight $\eta$ in semantic–spatial affinity estimation under confidence gating, and the calibration-signal weight $\alpha$ during token calibration. The optimal values are $\lambda=0.5$, $\eta=0.45$, $\alpha=0.15$. Further experimental details are provided in Supplementary Sec. D.

\begin{figure}[t]
\centering
\includegraphics[scale=0.48]{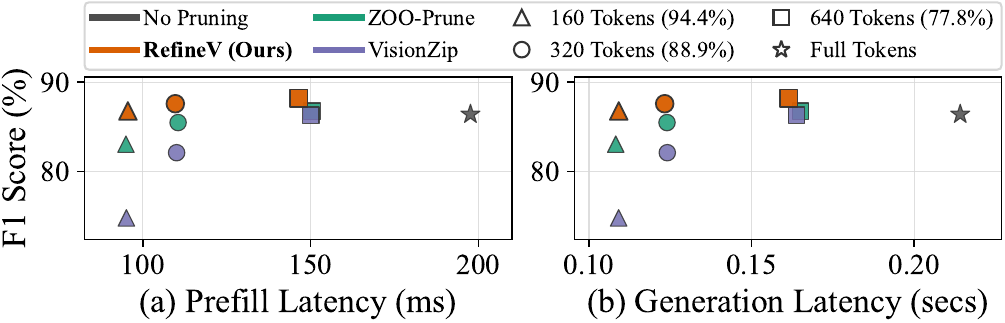} 
\caption{Inference efficiency comparison between RefineV and competing methods on LLaVA-NeXT-7B across different pruning ratios on POPE. }
\vspace{-3pt}
\label{figure4}
\end{figure}

\paragraph{Inference Efficiency.}
To evaluate efficiency, Figure \ref{figure4} compares the prefill and reported generation latency of RefineV with the selection method ZOO-Prune \cite{kim2026zoo} and the compression method VisionZip \cite{yang2025visionzip}. Prefilling is dominated by visual-token processing and is critical for pruning-based acceleration. RefineV maintains low latency while improving the accuracy--efficiency trade-off across token budgets.

Under unified inference and timing protocols on the same hardware, we measure prefilling time and end-to-end latency for three acceleration methods and the unpruned model. RefineV achieves up to a $2.30\times$ speedup over unpruned inference across the evaluated settings. Its calibration adds only lightweight tensor operations without additional LLM  forward passes; therefore, the associated overhead remains small and does not offset the acceleration gains. Appendix.C provides the corresponding evaluation details.

\paragraph{Qualitative Visualization}

As shown in Figure~\ref{figure7}, we present representative results on VQA$^T$, GQA, and VQA$^{v2}$ across three backbones. The examples illustrate how selection or compression may remove or mix critical visual evidence. 
PRCA-only results further show that retaining robust anchors does not always preserve sufficient information for downstream reasoning. By selectively incorporating reliably attributed residual signals, RefineV refines these anchors and improves predictions in the illustrated cases, supporting the role of semantic refinement beyond token selection alone.

\section{Conclusion}

Long visual token sequences make VLM inference costly, while token reduction may discard, dilute, or misassign critical visual cues. We present RefineV, a training-free framework that couples visual token reduction with semantic refinement before reasoning. PRCA selects compact anchors with stable model-side influence across perturbation directions, while CGTC filters residual signals by attribution reliability before refining their representations without increasing the token budget. Extensive evaluations across diverse VLM architectures show improved normalized performance under fixed budgets, together with reduced prefilling time and end-to-end latency. These results highlight the value of refining compact representations rather than treating token reduction as the end of visual processing.

\paragraph{Limitations}
RefineV remains sensitive to visual evidence that is weakly encoded, ambiguous, or severely degraded. Calibration refines and redistributes information already present in visual tokens but does not fully recover evidence lost during encoding. Future work will extend semantic refinement to visual encoding.

\section{Acknowledgments}
This work was supported by the National Natural Science Foundation of China (NSFC) under Grant No. 62441235.

\input{main.bbl}
\clearpage
\appendix

\twocolumn[
    \centering
    \vspace*{0.45in}
    {\LARGE\bfseries Supplementary Material\par}
    \vspace{0.58in}
]

\input{Supplementary_material}

\end{document}

%% file: Supplementary_material.tex
\setcounter{table}{0}
\renewcommand{\thetable}{S\arabic{table}}

\setcounter{figure}{0}
\renewcommand{\thefigure}{S\arabic{figure}}

This material provides supplementary discussions of the motivation behind our work, details of the TFLOPs calculation, additional information on experimental configurations and evaluation results, further visualizations and analyses of our method across different backbones and tasks. This material complements the main paper and provides a more comprehensive understanding of our method and experimental findings. The remainder of the supplementary material is organized as follows:
\begin{itemize}
    \item \textbf{Section A}: Detailed Discussion of the Motivation and Computation of Semantic Retention Presented in the introduction of the Main Paper.
    \item \textbf{Section B}: Theoretical Analysis of the TFLOPs Calculation Reported in Table 1 of the main paper.
    \item \textbf{Section C}: Implementation Details and Supplementary Results of the Comparative Experiments.
    \item \textbf{Section D}: Detailed Results and Analysis of the Ablation Studies and Hyperparameter Analysis.
    \item \textbf{Section E}: Additional Visualization Results and Corresponding Analysis.
    \item \textbf{Section F}: Discussion and Future Work.
\end{itemize}


\section{A. Detailed Discussion of the Motivation}
Fixed-budget visual token reduction aims to reduce visual tokens while preserving the information required for multimodal reasoning. Existing methods mainly rely on token selection, token merging and compression. However, satisfying a token budget does not guarantee semantic consistency with the full visual sequence. Task-relevant tokens may be pruned, semantically incompatible regions may be fused, and foreground evidence may be diluted by dominant background features. We refer to this discrepancy as \textbf{semantic drift}. The objective should therefore be not merely to retain fewer tokens, but to preserve the semantics required by the current question, which is crucial for model reasoning.

The upper panel of Figure~1 in the main paper illustrates these failure modes. The full sequence jointly represents the dog, cat, grass, and store. Selection-based reduction may remove less salient contextual tokens, causing the ``store'' semantics to disappear. Merging or compression may instead combine visually similar yet semantically different regions. For example, cat features may be mixed with road or background features, weakening the evidence for ``cat.'' The retained tokens may remain visually plausible, but their combined representation no longer covers the complete scene semantics.

To quantify this effect, the lower panel of Figure~1 in the main paper reports \textbf{Calibrated Semantic Retention (CSR)}, abbreviated as Semantic Retention in the bar chart. CSR is a question-conditioned representation-coverage metric that estimates how much source information remains recoverable from the reduced visual sequence. For each image--question pair, let $V=\{v_i\}_{i=1}^{N}$, $Q=\{t_l\}_{l=1}^{T}$, and $V_S=\{v_j\}_{j=1}^{K}$ denote the full visual tokens, non-special question-token embeddings, and reduced visual tokens, respectively. Both \(V\) and \(V_S\) are mapped through the same frozen multimodal projector, while \(Q\) is obtained from the frozen LLM embedding layer. We first measure the relevance of each source token to the question as follows:

\begin{equation}
r_i=
\max_{1\leq l\leq T}
\left[\bar v_i^{\top}\bar t_l\right]_{+},
\qquad
[x]_{+}=\max(x,0).
\label{eq:csr_relevance}
\end{equation}

The resulting relevance scores are normalized over the full visual sequence as follows:

\begin{equation}
w_i=
\begin{cases}
\dfrac{r_i}{\sum_{\ell=1}^{N}r_\ell},
& \sum_{\ell=1}^{N}r_\ell>0,\\[6pt]
\dfrac{1}{N},
& \text{otherwise}.
\end{cases}
\label{eq:csr_weight}
\end{equation}

The uniform fallback avoids an undefined score when none of the visual tokens has positive similarity to the question. Importantly, $w_i$ is computed once from the same uncompressed reference sequence and then reused for every reduction method. For each source token, we next measure whether its representation remains covered by any reduced token:

\begin{equation}
c_i=
\max_{1\leq j\leq K}
\left[\bar v_i^{\top}\bar v_j\right]_{+}.
\label{eq:csr_coverage}
\end{equation}

Here, $c_i\in[0,1]$ represents source-to-output coverage: a high value indicates that the semantics carried by $v_i$ still have a compatible carrier in the reduced sequence, irrespective of whether that carrier was produced by selection, merging, or residual calibration. CSR is then defined as the question-weighted coverage of the full source sequence:

\begin{equation}
\operatorname{CSR}
=
\sum_{i=1}^{N}w_i c_i.
\label{eq:csr_final}
\end{equation}

The bar chart reports $100\times\operatorname{CSR}$, averaged over the same 1,000 evaluated image-question pairs with the backbone of LLaVA-1.5-7B across GQA, MME, SQA and VQA$^T$. All methods are evaluated at the same output-token budget and with the same reference tokens, representation layer, image-question samples, and scoring rule; no method-specific retention mask, confidence gate, merge assignment, or perturbation score enters the calculation. CSR is used only as an auxiliary representation-level diagnostic rather than a causal measure of answer preservation.

Figure~2 in the main paper further contrasts conventional reduction with visual token calibration. Selection may preserve insufficient evidence and produce ``dog'', whereas compression may overmix foreground and background information and produce ``bird''. RefineV instead identifies robust \textbf{Calib-Anchors}, filters reliable residual signals, and routes them to compatible anchors. It therefore recovers useful semantics while maintaining the fixed output budget, providing a more reliable visual basis for predicting ``horse''.

\section{B. Theoretical Derivation of TFLOPs}

We estimate the single-sample computational cost of RefineV by decomposing inference into three stages: robust calibration-anchor selection, residual-signal calibration and fusion, and LLM prefill after token reduction. The visual encoder is excluded because it processes the complete image before token reduction and therefore introduces the same overhead for all compared methods. Let $N$ and $K$ denote the numbers of original and retained visual tokens, respectively, and let $T$ denote the number of text tokens. The number of discarded tokens $N_d$, retention ratio $\rho$, and sequence length $W$ entering the LLM are defined as:
\begin{equation}
N_d=N-K,\qquad
\rho=\frac{K}{N},\qquad
W=T+K.
\label{eq:basic_definitions}
\end{equation}

In PRCA, RefineV estimates the stable model-side influence of each visual token using $m$ symmetric perturbation directions and jointly considers token diversity during anchor selection. Its computational cost is approximated by
\begin{equation}
\mathcal{F}_{\mathrm{score}}
=
2mF_M(N)+O(mNH+N^2d_v+KN),
\label{eq:score_flops}
\end{equation}
where $d_v$ and $H$ are the dimension of the visual characteristics and the hidden dimension of LLM, $F_M(N)$ denotes the FLOPs of the multimodal projector in the actual backbone when processing $N$ tokens. The first term measures positive and negative perturbation responses through the projector, assuming that one multiply-add operation contributes two FLOPs. The remaining terms correspond to the estimation of the response-magnitude, the computation of pairwise diversity, and the iterative selection of calibration anchors $K$.

CGTC further examines the $N_d$ unselected tokens and extracts reliable complementary signals. Its cost is:
\begin{equation}
\mathcal{F}_{\mathrm{res}}
=
O(N_dKd_v+N_dK).
\label{eq:residual_flops}
\end{equation}
Here, $N_dKd_v$ measures the affinity between residual tokens and calibration anchors, while $N_dK$ represents scalar-level operations in gated mechanisms, weight normalization, and residual signal weighted fusion. The final term covers confidence gating, weight normalization, and weighted signal fusion. The complete visual-token
processing cost is:
\begin{equation}
\mathcal{F}_{\mathrm{RefineV\!-\!select}}
=
\mathcal{F}_{\mathrm{score}}
+
\mathcal{F}_{\mathrm{res}}.
\label{eq:right_select_flops}
\end{equation}

After calibration, the resulting $W$-token sequence is fed into the language model. For one decoder-only Transformer layer, the prefill cost is estimated as:
\begin{equation}
\mathcal{F}_{\mathrm{layer}}
=
8WH^2+6WHI+4W^2H,
\label{eq:layer_flops}
\end{equation}
where $I$ is the FFN intermediate dimension. The three terms respectively represent the Q/K/V/output projections, the up, gate, down projections in the gated FFN, and the two matrix
multiplications in self-attention. For an LLM containing $L$ Transformer layers, the total prefill cost becomes
\begin{equation}
\mathcal{F}_{\mathrm{prefill}}
=
L\left(8WH^2+6WHI+4W^2H\right).
\label{eq:prefill_flops}
\end{equation}

Finally, the total computational cost and its TFLOPs representation are given by
\begin{equation}
\begin{gathered}
\mathcal{F}_{\mathrm{total}}^{\mathrm{RefineV}}
=
\mathcal{F}_{\mathrm{RefineV\!-\!select}}
+
\mathcal{F}_{\mathrm{prefill}},
\\[2pt]
\mathrm{TFLOPs}
=
\frac{
\mathcal{F}_{\mathrm{total}}^{\mathrm{RefineV}}
}{
10^{12}
}.
\end{gathered}
\label{eq:total_tflops}
\end{equation}

This formulation separates the calibration overhead from the token-budget-dependent LLM prefill cost, enabling a consistent comparison across different retained-token settings. The comparison of TFLOPs between our method and the baseline is presented in Table \ref{tab:tflops}.

\begin{table}[t]
\centering
\fontsize{10}{12}\selectfont
\setlength{\tabcolsep}{3.8pt}
\renewcommand{\arraystretch}{1.08}

\begin{tabular}{l|cccc}
\specialrule{0.8pt}{0pt}{0pt}
\textbf{Backbone} 
& \textbf{method} & \textbf{Rate} & \textbf{Tokens} & \textbf{TFLOPs} \\
\specialrule{0.8pt}{0pt}{0pt}

\hline
LLaVA-1.5-7B 
& baseline & 0 & 576 & 8.54 \\

LLaVA-1.5-7B 
& RefineV & 66.7\% & 192 & 3.47 \\

LLaVA-1.5-7B 
& RefineV & 77.8\% & 128 & 3.02 \\

LLaVA-1.5-7B 
& RefineV & 88.9\% & 64 & 2.18 \\

\hline
LLaVA-NeXT-7B 
& baseline & 0 & 2880 & 42.7 \\

LLaVA-NeXT-7B 
& RefineV & 77.8\% & 640 & 12.5 \\

LLaVA-NeXT-7B 
& RefineV & 88.9\% & 320 & 8.2 \\

LLaVA-NeXT-7B 
& RefineV & 94.4\% & 160 & 6.1 \\

\hline
Qwen2.5-VL-7B 
& baseline & 0 & 1110 & 17.3 \\

Qwen2.5-VL-7B 
& RefineV & 80.0\% & 222 & 5.4 \\

Qwen2.5-VL-7B 
& RefineV & 90.0\% & 111 & 3.8 \\

\specialrule{0.8pt}{0pt}{0pt}
\end{tabular}

{\fontsize{10}{12}\selectfont
\caption{TFLOPs comparison between RefineV and the baseline across three backbones. Qwen2.5-VL-7B processes images at variable resolutions, its TFLOPs are calculated using a representative sample, which is encoded 1,110 tokens.}
\vspace{-10pt}
\label{tab:tflops}
}

\end{table}

\begin{figure*}[t]
\centering
\includegraphics[scale=0.77]{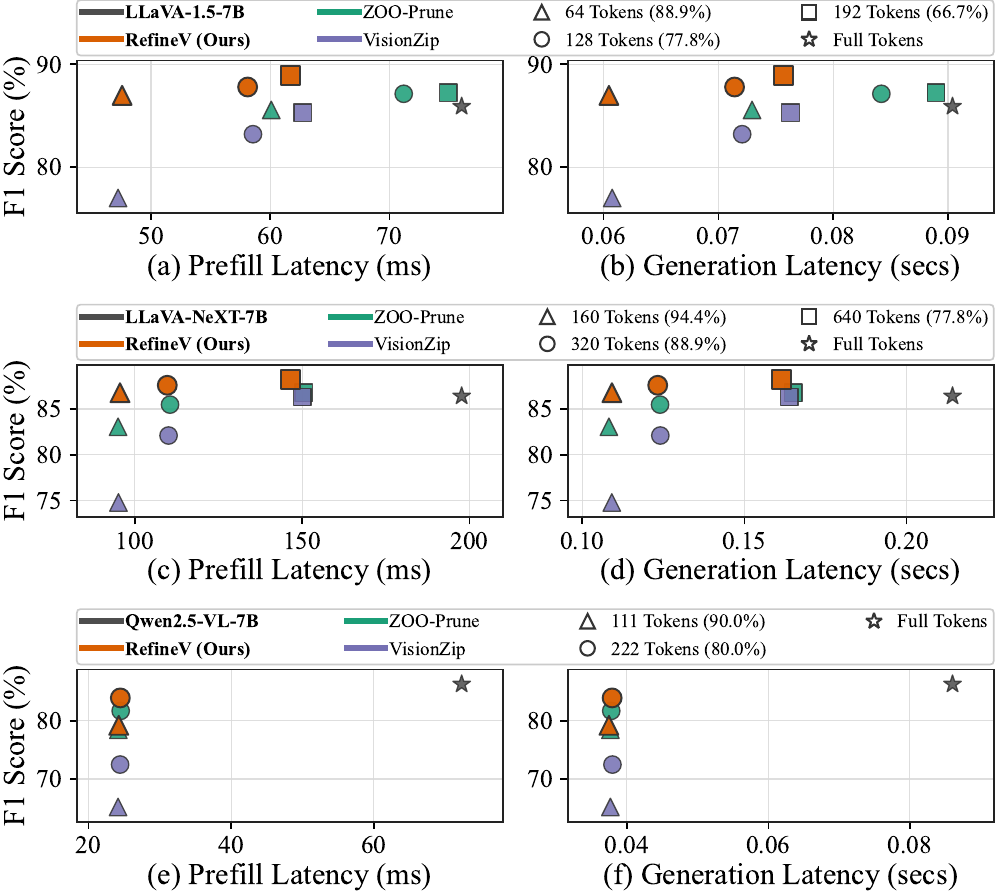} 
\caption{Inference efficiency comparison between RefineV and competing methods on LLaVA-1.5-7B, LLaVA-NeXT-7B and Qwen2.5-VL-7B across different pruning ratios on POPE with all 9,000 samples. }
\label{figures1}
\end{figure*}

\section{C. Experiment Setup}
\subsection{Model Settings}

We evaluate RefineV on three backbones: LLaVA-1.5-7B, LLaVA-NeXT-7B, and Qwen2.5-VL-7B. Both LLaVA-based models employ CLIP as the vision encoder and apply their default image preprocessing pipelines before visual encoding. Specifically, LLaVA-1.5-7B resizes each input image to $336 \times 336$, which is subsequently encoded into 576 visual tokens. LLaVA-NeXT-7B is designed to accommodate higher-resolution inputs and processes each image at a resolution of
$672 \times 672$, producing 2,880 visual tokens after CLIP encoding.

For the Qwen series, we adopt Qwen2.5-VL-7B, which is built upon the Qwen2.5-7B language model and employs a dynamic-resolution ViT encoder equipped with window attention. Consequently, the number of visual tokens varies according to the resolution of each input image. To ensure a fair comparison, we retain the original image preprocessing pipeline of each backbone without modification. All experiments are conducted on the same server under consistent hardware and software settings.

\subsection{Implementation Details}
All experiments are conducted on a single NVIDIA GeForce RTX 5090 D GPU with batch size 1 and four random seeds: 10, 20, 30, and 42. We report the mean performance across these runs. RefineV is entirely training-free and does not rely on attention scores for token selection or calibration. Its main computation is concentrated in the perturbation operations
performed around the lightweight projector layer. Consequently, the method requires neither backbone-specific retraining nor sample-specific optimization. Although the absolute computational cost varies with the number of input visual tokens, the overall calibration procedure can be applied directly to different backbones. Moreover, because most intermediate quantities required for calibration-signal selection have already been obtained during calibration-anchor construction, the additional overhead introduced by CGTC is relatively small. As shown in Figure \ref{figures1}, RefineV achieves a favorable performance--efficiency trade-off across all three backbones compared with the baseline and other conventional token reduction methods.

To provide a comprehensive evaluation under different token budgets, we adopt three pruning ratios of $66.7\%$, $77.8\%$ and $88.9\%$ for LLaVA-1.5-7B; $77.8\%$, $88.9\%$ and $94.4\%$ for LLaVA-NeXT-7B; and $80\%$ and $90\%$ for Qwen2.5-VL-7B. We set hyperparameter $h=5\times10^{-4}$, $m=64$, $\lambda=0.5$, $\eta=0.45$, $\theta_s=0.4$, $\theta_c=0.9$, $\alpha=0.15$, $\sigma_p=0.2$, $\tau_g=0.07$, $\tau_c=0.1$, $\tau_s=0.07$ and $\epsilon=1e-8$. Overall, the ablation studies demonstrate that RefineV remains robust across a broad range of hyperparameter settings. All evaluations are conducted using the  LMMs-Eval framework, following the official evaluation protocols and metrics for each benchmark.

\begin{table*}[t]
\centering
\fontsize{9}{10}\selectfont
\setlength{\tabcolsep}{7pt}
\renewcommand{\arraystretch}{1.08}

\begin{tabular}{l|ccccccccc|c}
\specialrule{0.8pt}{0pt}{0pt}
\textbf{Method} 
& \makecell{\textbf{GQA}}
& \makecell{\textbf{MMB}}
& \makecell{\textbf{MME}}
& \makecell{\textbf{POPE}}
& \makecell{\textbf{SQA}}
& \makecell{\textbf{VQA}$^{\mathbf{v2}}$}
& \makecell{\textbf{VQA}$^{\mathbf{T}}$}
& \makecell{\textbf{MMMU}}
& \makecell{\textbf{SEED}}
& \makecell{\textbf{Avg.}} \\
\specialrule{0.8pt}{0pt}{0pt}

\rowcolor{gray!20}
\multicolumn{11}{c}{\textit{Total 576 Tokens}} \\
\hline
LLaVA-1.5-7B 
& 61.90 & 64.70 & 1862.00 & 85.90 & 69.50 & 78.50 & 58.20 & 36.30 & 58.60 & 100\% \\
\hline

\rowcolor{gray!20}
\multicolumn{11}{c}{\textit{Retain 192 Tokens} \textcolor{green!70!black}{$\downarrow$ 66.7\%}} \\
\hline
ToMe (ICLR 2023) 
& 54.30 & 60.50 & 1563.00 & 72.40 & 65.20 & 68.00 & 52.10 & - & - & - \\

FastV (ECCV 2024) 
& 52.70 & 61.20 & 1612.00 & 64.80 & 67.30 & 67.10 & 52.50 & 34.30 & 57.10 & 89.6\% \\

HiRED (AAAI 2025) 
& 58.70 & 62.80 & 1737.00 & 82.80 & 68.40 & 74.90 & 47.40 & - & - & - \\

SparseVLM (ICML 2025) 
& 57.60 & 62.50 & 1721.00 & 83.60 & 69.10 & 75.60 & 56.10 & 33.80 & 55.80 & 95.5\% \\

PyramidDrop(CVPR 2025) 
& 57.10 & 63.20 & 1766.00 & 82.30 & 68.80 & 75.10 & 56.10 & - & - & - \\

VisionZip (CVPR 2025) 
& 59.30 & 63.00 & 1782.00 & 85.30 & 68.90 & 76.80 & 57.30 & 36.60 & 56.40 & 97.9\% \\

DivPrune (CVPR 2025) 
& 59.97 & 62.54 & 1762.30 & 87.00 & 68.91 & 76.87 & 56.97 & 35.44 & 58.71 & 98.0\% \\

FiCoCO-V (AAAI 2026) 
& 58.51 & 61.74 & 1755.61 & 81.47 & 67.42 & 74.22 & 55.50 & 36.11 & 57.95 & 96.0\% \\

V$^2$Drop (CVPR 2026) 
& 58.60 & 62.37 & 1789.76 & 85.10 & 69.30 & 74.92 & 55.62 & 35.87 & 56.74 & 97.0\% \\

ZOO-Prune (CVPR 2026) 
& 60.03 & 62.57 & 1781.66 & 87.24 & 69.01 & 75.35 & 57.10 & 35.44 & 58.78 & 98.0\% \\

\textbf{\textit{RefineV} (Ours)} 
& \textbf{60.57} & \textbf{63.41} & \textbf{1809.47} & \textbf{88.94} & \textbf{70.24} & \textbf{77.42} & \textbf{58.01} & \textbf{37.33} & \textbf{58.94} 
& \textcolor{red!80!black}{\textbf{99.9\%}} \\
\hline

\rowcolor{gray!20}
\multicolumn{11}{c}{\textit{Retain 128 Tokens} \textcolor{green!70!black}{$\downarrow$ 77.8\%}} \\
\hline
ToMe (ICLR 2023) 
& 52.40 & 53.30 & 1343.00 & 62.80 & 59.60 & 63.00 & 49.10 & - & - & - \\

FastV (ECCV 2024) 
& 49.60 & 56.10 & 1490.00 & 59.60 & 61.80 & 61.80 & 50.60 & 34.90 & 55.90 & 84.7\% \\

HiRED (AAAI 2025) 
& 57.20 & 61.50 & 1490.00 & 59.60 & 60.20 & 61.80 & 50.60 & - & - & - \\

SparseVLM (ICML 2025) 
& 56.00 & 60.00 & 1696.00 & 80.50 & 67.10 & 73.80 & 54.90 & 33.80 & 53.40 & 93.0\% \\

PyramidDrop(CVPR 2025) 
& 56.00 & 61.00 & 1644.00 & 82.30 & 68.30 & 72.90 & 55.10 & - & - & - \\

VisionZip (CVPR 2025) 
& 57.60 & 62.00 & \textbf{1761.70} & 83.20 & 68.90 & 75.60 & 56.80 & \textbf{37.90} & 54.90 & 96.8\% \\

DivPrune (CVPR 2025) 
& 59.25 & 62.03 & 1718.22 & 86.72 & 68.96 & 75.96 & 56.06 & 35.56 & 56.98 & 96.9\% \\

FiCoCO-V (AAAI 2026) 
& 57.59 & 60.34 & 1718.06 & 82.24 & 68.73 & 72.97 & 54.10 & 35.44 & 54.98 & 94.5\% \\

V$^2$Drop (CVPR 2026) 
& 56.19 & 59.28 & 1712.59 & 80.90 & 68.83 & 73.11 & 53.80 & 35.12 & 55.82 & 93.9\% \\

ZOO-Prune (CVPR 2026) 
& 59.43 & 61.86 & 1741.23 & 87.13 & 68.86 & 74.55 & 56.79 & 35.67 & 57.39 & 97.2\% \\

\textbf{\textit{RefineV} (Ours)} 
& \textbf{60.15} & \textbf{62.11} & 1731.46 & \textbf{87.79} & \textbf{69.89} & \textbf{76.73} & \textbf{57.96} & 36.22 & \textbf{57.73} 
& \textcolor{red!80!black}{\textbf{98.3\%}} \\
\hline

\rowcolor{gray!20}
\multicolumn{11}{c}{\textit{Retain 64 Tokens} \textcolor{green!70!black}{$\downarrow$ 88.9\%}} \\
\hline
ToMe (ICLR 2023) 
& 48.60 & 43.70 & 1138.00 & 52.50 & 50.00 & 57.10 & 45.30 & - & - & - \\

FastV (ECCV 2024) 
& 46.10 & 48.00 & 1256.00 & 48.00 & 51.10 & 55.00 & 47.80 & 34.00 & 51.90 & 75.5\% \\

HiRED (AAAI 2025) 
& 54.60 & 60.20 & 1599.00 & 73.60 & 68.20 & 69.70 & 44.20 & - & - & - \\

SparseVLM (ICML 2025) 
& 52.70 & 56.20 & 1505.00 & 75.10 & 62.20 & 68.20 & 51.80 & 32.70 & 51.10 & 87.0\% \\

PyramidDrop(CVPR 2025) 
& 41.90 & 33.30 & 1092.00 & 55.90 & 68.60 & 69.20 & 45.90 & - & - & - \\

VisionZip (CVPR 2025) 
& 55.10 & 60.10 & \textbf{1690.00} & 77.00 & 69.00 & 72.40 & 55.50 & 35.20 & 52.20 & 92.8\% \\

DivPrune (CVPR 2025) 
& 57.78 & 59.28 & 1674.40 & 85.56 & 68.17 & 74.11 & 54.69 & 35.56 & 55.13 & 94.8\% \\

FiCoCO-V (AAAI 2026) 
& 53.43 & 55.09 & 1653.67 & 75.29 & 67.92 & 71.63 & 51.88 & 34.22 & 53.14 & 90.1\% \\

V$^2$Drop (CVPR 2026) 
& 53.52 & 55.21 & 1641.23 & 75.14 & 68.92 & 71.03 & 51.83 & 34.29 & 53.11 & 90.1\% \\

ZOO-Prune (CVPR 2026) 
& 58.36 & 60.22 & 1626.80 & 85.56 & 68.17 & 72.97 & 55.35 & 35.21 & \textbf{55.84} & 94.8\% \\

\textbf{\textit{RefineV} (Ours)} 
& \textbf{59.12} & \textbf{60.58} & 1664.51 & \textbf{86.98} & \textbf{69.80} & \textbf{75.81} & \textbf{56.32} & \textbf{36.11} & 55.78 
& \textcolor{red!80!black}{\textbf{96.5\%}} \\

\specialrule{0.8pt}{0pt}{0pt}
\end{tabular}

{\fontsize{9}{9}\selectfont
\caption{Complete performance comparison on \textbf{LLaVA-1.5-7B}.}
\label{tab:supplement_llava15_7b}
}
\end{table*}

\subsection{Evaluation Details and Benchmarks}
We conduct a comprehensive evaluation of \textbf{RefineV} on nine widely used vision-language benchmarks, covering four complementary capability dimensions: visual question answering, advanced multimodal reasoning, object hallucination evaluation, and comprehensive multimodal assessment. All experiments follow the official evaluation protocols and data splits of the corresponding benchmarks. We evaluate all methods under the same model configuration, visual-token budget, and decoding settings to ensure a fair comparison. All evaluations are conducted in a single-model, zero-shot setting without task-specific fine-tuning.

For evaluation metrics, we report Accuracy (Acc.) for VQA$^{v2}$, GQA, ScienceQA-IMG, TextVQA, MMBench, MMMU and SEED-Bench. For POPE, we report the F1-score, which balances precision and recall in object-existence prediction. For MME, we report the sum of the Perception and Cognition scores, denoted as P+C. In addition to the individual benchmark results, we report a normalized average score to provide an overall comparison across benchmarks with different numerical scales. Specifically, the score on each benchmark is normalized by the performance of the corresponding unpruned model and the resulting ratios are averaged across all benchmarks:
\begin{equation}
    \mathrm{Avg.}
    =
    \frac{1}{N}
    \sum_{i=1}^{N}
    \frac{s_i}{s_i^{\mathrm{full}}}
    \times 100\%,
    \label{eq:normalized_average}
\end{equation}
where $s_i$ denotes the model performance on the $i$-th benchmark, while $s_i^{\mathrm{full}}$ denotes the performance of the corresponding unpruned model and $N$ is the number of diverse benchmarks evaluated.

\begin{table*}[t]
\centering
\fontsize{10}{12}\selectfont
\setlength{\tabcolsep}{5.2pt}
\renewcommand{\arraystretch}{1.08}

\begin{tabular}{l|ccccccccc|c}
\specialrule{0.8pt}{0pt}{0pt}
\textbf{Method} 
& \makecell{\textbf{GQA}}
& \makecell{\textbf{MMB}}
& \makecell{\textbf{MME}}
& \makecell{\textbf{POPE}}
& \makecell{\textbf{SQA}}
& \makecell{\textbf{VQA}$^{\mathbf{v2}}$}
& \makecell{\textbf{VQA}$^{\mathbf{T}}$}
& \makecell{\textbf{MMMU}}
& \makecell{\textbf{SEED}}
& \makecell{\textbf{Avg.}} \\
\specialrule{0.8pt}{0pt}{0pt}

\rowcolor{gray!20}
\multicolumn{11}{c}{\textit{Total 2880 Tokens}} \\
\hline
LLaVA-NeXT-7B 
& 64.20 & 67.90 & 1842.00 & 86.40 & 70.20 & 80.10 & 61.30 & 35.10 & 70.20 & 100\% \\
\hline

\rowcolor{gray!20}
\multicolumn{11}{c}{\textit{Retain 640 Tokens} \textcolor{green!70!black}{$\downarrow$ 77.8\%}} \\
\hline
SparseVLM (ICML 2025) 
& 60.30 & 65.70 & 1772.00 & - & 67.70 & 77.10 & 57.80 & 34.60 & - & - \\

VisionZip (CVPR 2025) 
& 61.30 & \textbf{66.30} & 1787.00 & 86.30 & 68.10 & \textbf{79.10} & 60.20 & 34.70 & 66.70 & 97.5\% \\

DivPrune (CVPR 2025) 
& 61.58 & 65.38 & 1773.04 & 85.51 & 67.82 & 78.94 & 55.41 & 36.89 & 67.56 & 97.1\% \\

FiCoCO-V (AAAI 2026) 
& 60.40 & 64.18 & 1780.22 & 86.04 & 67.94 & 75.77 & 54.89 & 35.91 & 65.77 & 95.8\% \\

ZOO-Prune (CVPR 2026) 
& 62.19 & 65.10 & 1802.47 & 86.75 & 68.02 & 78.13 & 57.98 & 36.82 & 67.59 & 97.9\% \\

\textbf{\textit{RefineV }(Ours)} 
& \textbf{63.87} & 65.98 & \textbf{1831.14} & \textbf{88.19} & \textbf{69.41} & 78.02 & \textbf{58.39} & \textbf{37.89} & \textbf{67.92} 
& \textcolor{red!80!black}{\textbf{99.4\%}} \\
\hline

\rowcolor{gray!20}
\multicolumn{11}{c}{\textit{Retain 320 Tokens} \textcolor{green!70!black}{$\downarrow$ 88.9\%}} \\
\hline
SparseVLM (ICML 2025) 
& 57.70 & 64.30 & 1694.00 & - & 67.30 & 73.40 & 55.90 & 34.40 & - & - \\

VisionZip (CVPR 2025) 
& 59.30 & 63.10 & 1702.00 & 82.10 & 67.30 & 76.20 & \textbf{58.90} & 35.30 & 63.40 & 94.5\% \\

DivPrune (CVPR 2025) 
& 59.63 & 63.66 & 1731.04 & 83.47 & 67.82 & 76.64 & 53.84 & 37.11 & 65.35 & 95.1\% \\

FiCoCO-V (AAAI 2026) 
& 59.26 & 63.30 & 1721.74 & 85.18 & 67.51 & 74.08 & 54.67 & 34.33 & 64.51 & 93.9\% \\

ZOO-Prune (CVPR 2026) 
& 60.83 & 64.60 & 1787.68 & 85.47 & 67.08 & 76.04 & 57.26 & 36.78 & 66.18 & 96.5\% \\

\textbf{\textit{RefineV }(Ours)} 
& \textbf{61.98} & \textbf{64.92} & \textbf{1791.90} & \textbf{87.58} & \textbf{68.86} & \textbf{77.76} & 57.67 & \textbf{37.33} & \textbf{66.28} 
& \textcolor{red!80!black}{\textbf{97.9\%}} \\
\hline

\rowcolor{gray!20}
\multicolumn{11}{c}{\textit{Retain 160 Tokens} \textcolor{green!70!black}{$\downarrow$ 94.4\%}} \\
\hline
SparseVLM (ICML 2025) 
& 51.20 & 63.10 & 1542.00 & - & 67.50 & 66.30 & 46.40 & 32.80 & - & - \\

VisionZip (CVPR 2025) 
& 55.50 & 60.10 & 1630.00 & 74.80 & 68.30 & 71.40 & \textbf{56.20} & 36.10 & 58.30 & 90.4\% \\

DivPrune (CVPR 2025) 
& 57.79 & 62.29 & 1658.25 & 79.36 & 68.02 & 73.92 & 52.42 & 36.44 & 62.54 & 92.4\% \\

FiCoCO-V (AAAI 2026) 
& 58.11 & 60.32 & 1657.81 & 84.27 & 68.01 & 73.22 & 55.10 & 33.98 & 63.06 & 92.4\% \\

ZOO-Prune (CVPR 2026) 
& 59.48 & 64.18 & 1706.08 & 83.05 & 67.72 & 74.07 & 55.42 & \textbf{36.67} & 64.04 & 94.5\% \\

\textbf{\textit{RefineV }(Ours)} 
& \textbf{59.97} & \textbf{64.47} & \textbf{1706.96} & \textbf{86.77} & \textbf{69.70} & \textbf{75.81} & 55.77 & 36.56 & \textbf{67.87}
& \textcolor{red!80!black}{\textbf{96.4\%}} \\

\specialrule{0.8pt}{0pt}{0pt}
\end{tabular}

{\fontsize{9}{10}\selectfont
\caption{Complete performance comparison on \textbf{LLaVA-NeXT-7B}.}
\label{tab:supplement_llava16_7b}
}

\end{table*}

\paragraph{Visual Question Answering.}
This category evaluates whether a model can understand visual content and answer questions grounded in the input image. We adopt four representative benchmarks:

\begin{itemize}
    \item \textbf{VQA$^{v2}$} is a general-purpose visual question answering benchmark containing open-ended questions about real-world images. Its balanced data construction reduces language priors and requires models to jointly understand visual content and natural-language questions.

    \item \textbf{GQA} focuses on compositional reasoning over  diverse scene graphs. It contains structured questions involving object attributes, spatial relations, multi-step visual reasoning, and so on.

    \item \textbf{ScienceQA-IMG} evaluates multimodal scientific reasoning using questions accompanied by images or diagrams. It requires both domain knowledge and the ability to interpret visual evidence.

    \item \textbf{TextVQA} assesses the ability to recognize and reason over textual information embedded in natural images. Successful performance therefore depends on both optical character recognition and contextual reasoning.
\end{itemize}

\paragraph{Advanced Multimodal Reasoning.}
Beyond conventional visual question answering, we evaluate the model's ability to perform fine-grained perception, knowledge-intensive reasoning, and complex cross-modal inference on three challenging benchmarks:

\begin{itemize}
    \item \textbf{MMBench} evaluates a broad range of perception and reasoning capabilities across multiple fine-grained skill dimensions, including attribute recognition, spatial understanding, logical reasoning, knowledge-based inference, and so on.

    \item \textbf{MMMU} is a multidisciplinary benchmark containing expert-level questions from more than 30 academic subjects, which are grouped into six major disciplines. Many questions involve complex diagrams, charts, tables, and domain-specific knowledge.

    \item \textbf{SEED-Bench} evaluates multimodal large language models through diverse visually grounded multiple-choice questions. It covers a wide range of perception and reasoning abilities and provides a unified assessment of general multimodal understanding.
\end{itemize}

\begin{table*}[t]
\centering
\fontsize{10}{12}\selectfont
\setlength{\tabcolsep}{7.5pt}
\renewcommand{\arraystretch}{1.08}

\begin{tabular}{l|cccccccc|c}
\specialrule{0.8pt}{0pt}{0pt}
\textbf{Method} 
& \textbf{GQA} 
& \textbf{MMB} 
& \textbf{MME} 
& \textbf{POPE} 
& \textbf{SQA}
& \textbf{VQA$^T$}
& \textbf{MMMU}
& \textbf{SEED}
& \textbf{Avg.-8} \\
\specialrule{0.8pt}{0pt}{0pt}

\rowcolor{gray!20}
\multicolumn{10}{c}{\textit{Full Tokens}} \\
\hline
Qwen2.5-VL-7B 
& 60.84 & 82.82 & 2310.00 & 86.30 & 88.30 & 77.69 & 50.22 & 74.21 & 100\% \\
\hline

\rowcolor{gray!20}
\multicolumn{10}{c}{\textit{Retain 20\% Tokens} \textcolor{green!70!black}{$\downarrow$ 80.0\%}} \\
\hline
VisionZip (CVPR 2025)
& 57.27 & 76.72 & 2221.00 & 81.67 & - & - & - & - & - \\

DivPrune (CVPR 2025)
& \textbf{60.05} & 75.87 & 2173.00 & 83.11 & - & - & - & - & - \\

FiCoCO-V (AAAI 2026)
& 55.28 & 76.47 & 2120.00 & 81.29 & - & - & - & - & - \\

ZOO-Prune (CVPR 2026)
& 55.92 & 77.41 & 2139.14 & 81.71 & 82.40 & 68.80 & 47.11 & 69.67 & 92.8\% \\

\textbf{\textit{RefineV} (Ours)} 
& 58.11 & \textbf{77.91} & \textbf{2274.28} & \textbf{83.94} & \textbf{84.43} & \textbf{70.85} & \textbf{47.68} & \textbf{73.16} & \textcolor{red!80!black}{\textbf{95.7\%}} \\
\hline

\rowcolor{gray!20}
\multicolumn{10}{c}{\textit{Retain 10\% Tokens} \textcolor{green!70!black}{$\downarrow$ 90.0\%}} \\
\hline
VisionZip (CVPR 2025)
& 54.09 & 72.16 & 1937.00 & 78.97 & - & - & - & - & - \\

DivPrune (CVPR 2025)
& \textbf{55.49} & 71.95 & 2004.00 & 79.05 & - & - & - & - & - \\

FiCoCO-V (AAAI 2026)
& 52.79 & 71.95 & 1967.00 & 77.61 & - & - & - & - & - \\

ZOO-Prune (CVPR 2026)
& 53.28 & 72.51 & 1961.89 & 78.46 & 79.92 & 61.99 & 45.33 & 66.86 & 87.7\% \\

\textbf{\textit{RefineV} (Ours)} 
& 54.45 & \textbf{73.37} & \textbf{2016.92} & \textbf{79.23} & \textbf{81.54} & \textbf{65.33} & \textbf{46.11} & \textbf{69.30} & \textcolor{red!80!black}{\textbf{89.9\%}} \\
\hline

\specialrule{0.8pt}{0pt}{0pt}
\end{tabular}

{\fontsize{10}{12}\selectfont
\caption{Complete performance comparison on \textbf{Qwen2.5-VL-7B}.}
\label{tab:supplement_qwen25_7b}
}

\end{table*}

\paragraph{Object Hallucination Evaluation.}
We employ \textbf{POPE} to evaluate object hallucination, a critical failure mode in which a model incorrectly claims that an absent object appears in an image. POPE formulates object-existence verification as binary questions and measures whether the model can distinguish objects that are present from those that are absent. We report the F1-score to jointly account for precision and recall, thereby providing a relatively balanced evaluation of hallucination resistance.

\paragraph{Comprehensive Multimodal Assessment.}
We further adopt \textbf{MME} to assess the model's overall multimodal capability. MME covers a broad collection of perception-oriented and cognition-oriented tasks, including object recognition, optical character recognition, spatial understanding, commonsense reasoning, and numerical reasoning. It reports separate scores for Perception and Cognition, and we use their sum, denoted as P+C, as the final evaluation metric.

\subsection{Comparative Experiment Analysis}
Unlike previous token reduction approaches that either select visual tokens according to different importance scores or compress them through token merging, RefineV formulates token reduction as a token calibration problem. We analyze its performance on three different backbones below.

\paragraph{Results on LLaVA-1.5-7B.}
As shown in Table \ref{tab:supplement_llava15_7b}, \textbf{RefineV} consistently outperforms both score-based token selection and token compression methods across all three token budgets. With 192, 128, and 64 retained tokens, it achieves normalized average performance of 99.9\%, 98.3\%, and 96.5\%, exceeding
the strongest prior method by 1.9, 1.1 and 1.7 percentage points, respectively. At 192 tokens, RefineV even surpasses the full-token model on POPE, SQA, MMMU, and SEED, while preserving nearly all performance on the remaining benchmarks. Its advantage becomes more pronounced under aggressive reduction: at 64 tokens, it obtains the best POPE, SQA, VQA$^{v2}$ and VQA$^T$ results, indicating that calibration recovers task-relevant information that conventional selection or compression may discard or dilute. Isolated gaps remain on MME and MMMU at 128 tokens, and on MME and SEED at 64 tokens. These cases mainly involve dense global evidence or fine-grained textual cues, yet the gaps are limited and RefineV still delivers the strongest overall performance--efficiency trade-off.

\paragraph{Results on LLaVA-NeXT-7B.}
As shown in Table \ref{tab:supplement_llava16_7b}, \textbf{RefineV} shows stable superiority under increasingly aggressive token reduction. It retains 99.4\%, 97.9\%, and 96.4\% of full-model performance with 640, 320 and 160 tokens, outperforming the strongest competing method by 1.5, 1.4 and 1.9 percentage points, respectively. At 640 tokens, it nearly matches the full model on MME and exceeds it on POPE and MMMU, demonstrating that token calibration can remove substantial redundancy without weakening and sometimes even improving reliability of reasoning. With 320 tokens, RefineV ranks first on eight of the nine reported benchmarks. In particular, its advantages on POPE and SEED remain clear at the most aggressive setting, showing strong robustness in hallucination control and general multimodal understanding. VQA$^T$ is the only consistent exception, where VisionZip performs slightly better, while VisionZip also leads marginally on VQA$^{v2}$ at 640 tokens. These small deficits may reflect the difficulty of preserving OCR-sensitive local details, but the substantially higher normalized averages confirm the overall advantage of the visual token calibration formulation.

\paragraph{Results on Qwen2.5-VL-7B.}
Unlike Table 2 in the main paper, Table \ref{tab:supplement_qwen25_7b} reports the performance of RefineV and the main competing methods across eight benchmarks. Consequently, the “Avg.” values differ slightly from those reported in the main paper. Nevertheless, \textbf{RefineV} consistently surpasses both ZOO-Prune, a score-based selection method, and FiCoCO-V, a token compression method. Retaining only 20\% and 10\% of visual tokens, it preserves 95.7\% and 89.9\% of the full-model performance, improving over ZOO-Prune by 2.9 and 2.2 percentage points in normalized average score. At the 20\% budget, RefineV remains superior on every metric reported by ZOO-Prune and FiCoCO-V. VQA$^T$ shows one of the most visible degradations relative to the full-token model among the accuracy-based benchmarks, reflecting its dependence on fine-grained OCR cues. Nevertheless, RefineV achieves its largest relative advantage on this benchmark, indicating that calibration better preserves and restores critical textual evidence under severe token reduction.

\begin{table*}[t]
\centering
\fontsize{10}{12}\selectfont
\setlength{\tabcolsep}{5pt}
\renewcommand{\arraystretch}{1.08}

\begin{tabular}{l|ccccccccc|c}
\specialrule{0.8pt}{0pt}{0pt}
\textbf{Module} 
& \makecell{\textbf{GQA}}
& \makecell{\textbf{MMB}}
& \makecell{\textbf{MME}}
& \makecell{\textbf{POPE}}
& \makecell{\textbf{SQA}}
& \makecell{\textbf{VQA}$^{\mathbf{v2}}$}
& \makecell{\textbf{VQA}$^{\mathbf{T}}$}
& \makecell{\textbf{MMMU}}
& \makecell{\textbf{SEED}}
& \makecell{\textbf{Avg.}} \\
\specialrule{0.8pt}{0pt}{0pt}

\rowcolor{gray!20}
\multicolumn{11}{c}{\textit{Total 576 Tokens}} \\
\hline
LLaVA-1.5-7B 
& 61.90 & 64.70 & 1862.00 & 85.90 & 69.50 & 78.50 & 58.20 & 36.30 & 58.60 & 100\% \\
\hline

\rowcolor{gray!20}
\multicolumn{11}{c}{\textit{Retain 192 Tokens} \textcolor{green!70!black}{$\downarrow$ 66.7\%}} \\
\hline
Only Sense-diversity
& 60.03 & 62.57 & 1781.66 & 87.24 & 69.01 & 75.35 & 57.10 & 35.44 & 58.78 & 98.0\% \\

Only Calib-Anchor
& 60.11 & 63.01 & 1784.31 & 87.78 & 69.80 & 77.01 & 57.19 & 36.56 & 58.89 & 99.0\% \\

Only Simple-Calib
& 60.18 & 62.98 & 1787.18 & 88.63 & 69.06 & 75.51 & 57.29 & 37.03 & 58.82 & 98.9\% \\

Only Confidence-Calib
& 60.52 & 63.23 & 1801.97 & 88.72 & 68.77 & 75.46 & 57.49 & 35.98 & 58.87 & 98.8\% \\

Calib-Anchor+Simple-Calib
& 60.21 & 63.34 & \textbf{1809.51} & 88.76 & 69.85 & 77.12 & 57.80 & 37.02 & 58.93 & 99.6\% \\

\textbf{\textit{Full RefineV }(Ours)} 
& \textbf{60.57} & \textbf{63.41} & 1809.47 & \textbf{88.94} & \textbf{70.24} & \textbf{77.42} & \textbf{58.01} & \textbf{37.33} & \textbf{58.94} 
& \textcolor{red!80!black}{\textbf{99.9\%}} \\
\hline

\rowcolor{gray!20}
\multicolumn{11}{c}{\textit{Retain 128 Tokens} \textcolor{green!70!black}{$\downarrow$ 77.8\%}} \\
\hline
Only Sense-diversity
& 59.43 & 61.86 & 1741.23 & 87.13 & 68.86 & 74.55 & 56.79 & 35.67 & 57.39 & 97.2\% \\

Only Calib-Anchor
& 59.72 & 61.36 & 1733.30 & 87.50 & 69.42 & \textbf{76.73} & 56.62 & 36.11 & 57.46 & 97.6\% \\

Only Simple-Calib
& 59.50 & 61.89 & 1719.87 & 87.59 & 68.91 & 74.81 & 56.85 & 36.11 & 57.67 & 97.4\% \\

Only Confidence-Calib
& 59.87 & 62.00 & \textbf{1747.59} & 87.58 & 68.52 & 74.76 & 56.96 & 35.89 & 57.55 & 97.5\% \\

Calib-Anchor+Simple-Calib
& 59.92 & 61.77 & 1733.59 & 87.47 & 69.72 & 76.21 & 57.53 & \textbf{36.22} & 57.58 & 97.9\% \\

\textbf{\textit{Full RefineV }(Ours)} 
& \textbf{60.15} & \textbf{62.11} & 1731.46 & \textbf{87.79} & \textbf{69.89} & \textbf{76.73} & \textbf{57.96} & \textbf{36.22} & \textbf{57.73} 
& \textcolor{red!80!black}{\textbf{98.3\%}} \\
\hline

\rowcolor{gray!20}
\multicolumn{11}{c}{\textit{Retain 64 Tokens} \textcolor{green!70!black}{$\downarrow$88.9\%}} \\
\hline
Only Sense-diversity
& 58.36 & 60.22 & 1626.80 & 85.56 & 68.17 & 72.97 & 55.35 & 35.21 & \textbf{55.84} & 94.8\% \\

Only Calib-Anchor
& 58.61 & 60.23 & 1631.64 & 86.28 & 68.71 & 75.61 & 56.26 & 35.44 & 55.46 & 95.6\% \\

Only Simple-Calib
& 58.51 & 60.54 & 1644.63 & 86.68 & 68.77 & 73.29 & 55.75 & 35.33 & 55.62 & 95.3\% \\

Only Confidence-Calib
& 58.95 & 60.57 & 1648.61 & 86.53 & 68.52 & 73.25 & 55.71 & 35.56 & 55.71 & 95.4\% \\

Calib-Anchor+Simple-Calib
& 58.98 & 60.57 & 1657.81 & 86.54 & 69.60 & 75.76 & 56.15 & 35.56 & 55.62 & 96.1\% \\

\textbf{\textit{Full RefineV }(Ours)} 
& \textbf{59.12} & \textbf{60.58} & \textbf{1664.51} & \textbf{86.98} & \textbf{69.80} & \textbf{75.81} & \textbf{56.32} & \textbf{36.11} & 55.78
& \textcolor{red!80!black}{\textbf{96.5\%}} \\

\specialrule{0.8pt}{0pt}{0pt}
\end{tabular}

{\fontsize{10}{12}\selectfont
\caption{Module ablation on LLaVA-1.5-7B.}
\label{tab:supplement_module_ablation}
}

\end{table*}

\section{D. Ablation Studies and Hyperparameter Analysis}
\subsection{Ablation Studies with Module}
As shown in Table \ref{tab:supplement_module_ablation}, we conduct more comprehensive module-wise ablation studies on LLaVA-1.5-7B under three token-reduction ratios, namely 66.7\%, 77.8\% and 88.9\%, to examine the contribution of each component in RefineV. The unpruned LLaVA-1.5-7B serves as the full-token reference. \textbf{Only Sense-Diversity} retains the original score-based selection strategy, which selects visual tokens according to sensitivity and diversity scores without applying token calibration. \textbf{Only Calib-Anchor} introduces PRCA to identify perturbation-robust calibration anchors, but does not perform subsequent information calibration. \textbf{Only Simple-Calib} applies the original selector and directly transfers residual information to the selected tokens without confidence gating. \textbf{Only Confidence-Calib} first selects tokens using the original strategy and then calibrates them through CGTC. \textbf{Calib-Anchor+Simple-Calib} combines PRCA-based anchor selection with ungated candidate-to-anchor calibration. Finally, \textbf{Full RefineV} denotes the complete RefineV framework integrating both PRCA and CGTC.

The results reveal a clear overall trend: performance improves as the calibration pipeline becomes more complete. Compared with Only Sense-Diversity, introducing PRCA alone increases the normalized average from 98.0\% to 99.0\%, from 97.2\% to 97.6\%, and from 94.8\% to 95.6\% under the three respective token budgets. Both ``Simple-Calib'' and ``Confidence-Calib'' also consistently improve upon the selection-only baseline, confirming that recovering residual information is beneficial even without robust anchor identification. Combining PRCA with ``Simple-Calib'' further raises the average performance to 99.6\%, 97.9\% and 96.1\%, demonstrating that robust anchors provide more reliable destinations for information transfer.

The complete model achieves the best normalized averages of 99.9\%, 98.3\% and 96.5\%. Relative to ``Calib-Anchor+Simple-Calib'', confidence-gated calibration yields additional gains of 0.3, 0.4 and 0.4 percentage points, indicating that indiscriminate information transfer is suboptimal and that CGTC effectively suppresses unreliable calibration signals. At 64 tokens, it remains best on eight of the nine benchmarks. The only exception is SEED, where Only Sense-Diversity is marginally higher by 0.06 points. Similarly, at 192 tokens, ``Calib-Anchor + Simple-Calib'' slightly outperforms the full model on MME. These isolated fluctuations suggest that benchmark-specific optima may occasionally differ, but they do not affect the consistent advantage of the complete framework. Together with the comparative results, the ablations verify that PRCA and CGTC are complementary and jointly enable RefineV to outperform previous token reduction methods.

\subsection{Hyperparameter Analysis}
\paragraph{Impact of $h$ and $m$.}
In the Perturbation-Robust Calibration Anchoring (PRCA) module, the perturbation step size and the number of perturbation directions are set to $h=5\times10^{-4}$ and $m=64$, respectively. The step size $h$ controls the magnitude of the symmetric perturbations applied to visual tokens before they enter the projector. An excessively large perturbation may move the token representation away from its local neighborhood, whereas an overly small perturbation may produce responses that are too weak to reliably characterize the projector's sensitivity. We therefore select an intermediate value that yields stable and distinguishable perturbation responses when selecting calibrated anchors.

The hyperparameter $m$ specifies the number of perturbation directions used to estimate the model-side influence of each visual token. We organize these perturbations into paired positive and negative directions, enabling a symmetric estimation of the projector response. Increasing $m$ generally produces a more reliable estimate by covering a broader set of local directions, but it also introduces additional computation. Considering both accuracy and efficiency, we set the number of perturbation directions $m=64$. 

\begin{table}[t]
\centering
\fontsize{9}{11}\selectfont
\setlength{\tabcolsep}{4pt}
\renewcommand{\arraystretch}{1.08}

\begin{tabular}{c|cccccc|c}
\specialrule{0.8pt}{0pt}{0pt}
\textbf{$\lambda$} 
& GQA & MMB & MME & POPE & SQA & VQA$^T$ & Avg. \\
\specialrule{0.8pt}{0pt}{0pt}

\rowcolor{gray!20}
\multicolumn{8}{c}{\textit{Total 576 Tokens}} \\
\hline
-
& 61.90 & 64.70 & 1862 & 85.90 & 69.50 & 58.20 & 100\% \\
\hline

\rowcolor{gray!20}
\multicolumn{8}{c}{\textit{Retain 192 Tokens} \textcolor{green!70!black}{$\downarrow$ 66.7\%}} \\
\hline

0.00
& 60.42 & 63.06 & 1786 & 87.68 & 69.51 & 57.88 & 98.8\% \\

0.25 
& 60.41 & 63.26 & 1798 & 87.68 & 69.94 & 57.90 & 99.0\% \\

0.50 
& \textbf{60.57} & \textbf{63.41} & \textbf{1809} & \textbf{88.94} & \textbf{70.24} & \textbf{58.01} & \textcolor{red!80!black}{\textbf{99.6\%}} \\

0.75 
& 60.42 & 63.06 & 1797 & 88.23 & 69.93 & 57.88 & 99.1\% \\

\hline

\rowcolor{gray!20}
\multicolumn{8}{c}{\textit{Retain 128 Tokens} \textcolor{green!70!black}{$\downarrow$ 77.8\%}} \\
\hline

0.00
& 60.04 & 61.94 & 1709 & 87.60 & 69.60 & 57.44 & 97.6\% \\

0.25 
& 60.14 & 61.94 & 1721 & 87.62 & 69.66 & 57.40 & 97.7\% \\

0.50 
& \textbf{60.15} & \textbf{62.11} & \textbf{1731} & \textbf{87.79} & \textbf{69.89} & \textbf{57.96} & \textcolor{red!80!black}{\textbf{98.1\%}} \\

0.75 
& 60.00 & 61.86 & 1719 & 87.58 & 69.60 & 57.44 & 97.6\% \\
\hline

\rowcolor{gray!20}
\multicolumn{8}{c}{\textit{Retain 64 Tokens} \textcolor{green!70!black}{$\downarrow$ 88.9\%}} \\
\hline

0.00
& 58.81 & 60.39 & 1652 & 86.51 & 69.06 & 56.10 & 95.6\% \\

0.25 
& 58.82 & \textbf{60.61} & 1651 & 86.53 & 69.11 & 56.14 & 95.7\% \\

0.50 
& \textbf{59.12} & 60.58 & \textbf{1665} & \textbf{86.98} & \textbf{69.80} & \textbf{56.32} & \textcolor{red!80!black}{\textbf{96.2\%}} \\

0.75 
& 58.82 & 60.31 & 1656 & 86.52 & 68.02 & 56.04 & 95.3\% \\
\hline

\specialrule{0.8pt}{0pt}{0pt}
\end{tabular}

{\fontsize{9}{10}\selectfont
\caption{Ablation study of $\lambda$ on LLaVA-1.5-7B.}
\label{tab:supplement_lambda_ablation}
}

\end{table}

\paragraph{Impact of $\lambda$.}
The robustness coefficient $\lambda$ controls the contribution of response variation across perturbation directions to anchor scoring. When $\lambda=0$, anchor selection degenerates into ranking tokens solely according to their mean response magnitude. However, a token may exhibit a strong response along only one direction while remaining nearly insensitive to the others. Such a token has a large response in a limited local direction but does not exert a stable influence on the model. Incorporating directional variation therefore helps PRCA prioritize tokens with consistently strong model-side effects. As shown in Table \ref{tab:supplement_lambda_ablation}, ablation results identify $\lambda=0.5$ as the optimal setting.

\begin{table}[t]
\centering
\fontsize{9}{11}\selectfont
\setlength{\tabcolsep}{4pt}
\renewcommand{\arraystretch}{1.08}

\begin{tabular}{c|cccccc|c}
\specialrule{0.8pt}{0pt}{0pt}
\textbf{$\eta$} 
& GQA & MMB & MME & POPE & SQA & VQA$^T$ & Avg. \\
\specialrule{0.8pt}{0pt}{0pt}

\rowcolor{gray!20}
\multicolumn{8}{c}{\textit{Total 576 Tokens}} \\
\hline
-
& 61.90 & 64.70 & 1862 & 85.90 & 69.50 & 58.20 & 100\% \\
\hline

\rowcolor{gray!20}
\multicolumn{8}{c}{\textit{Retain 192 Tokens} \textcolor{green!70!black}{$\downarrow$ 66.7\%}} \\
\hline

0.00
& 59.89 & 61.77 & 1698 & 87.41 & 68.96 & 57.36 & 97.2\% \\

0.15 
& 60.21 & 62.11 & 1763 & 88.13 & 69.51 & 57.61 & 98.3\% \\

0.30 
& 60.26 & 62.23 & 1772 & 88.48 & 69.80 & 57.89 & 98.6\% \\

0.45 
& \textbf{60.57} & \textbf{63.41} & \textbf{1809} & \textbf{88.94} & \textbf{70.24} & \textbf{58.01} & \textcolor{red!80!black}{\textbf{99.6\%}} \\

0.60 
& 60.27 & 62.54 & 1795 & 88.26 & 69.60 & 57.75 & 98.8\% \\

\hline

\rowcolor{gray!20}
\multicolumn{8}{c}{\textit{Retain 128 Tokens} \textcolor{green!70!black}{$\downarrow$ 77.8\%}} \\
\hline

0.00
& 59.39 & 60.65 & 1649 & 87.52 & 69.11 & 56.53 & 96.1\% \\

0.15 
& 59.55 & 62.00 & 1708 & 87.68 & 69.35 & 57.28 & 97.3\% \\

0.30 
& 59.92 & 61.83 & 1719 & 87.62 & 69.60 & 57.43 & 97.6\% \\

0.45 
& \textbf{60.15} & \textbf{62.11} & \textbf{1731} & \textbf{87.79} & \textbf{69.89} & \textbf{57.96} & \textcolor{red!80!black}{\textbf{98.1\%}} \\

0.60 
& 59.85 & 61.86 & 1724 & 87.51 & 69.40 & 57.22 & 97.5\% \\
\hline

\rowcolor{gray!20}
\multicolumn{8}{c}{\textit{Retain 64 Tokens} \textcolor{green!70!black}{$\downarrow$ 88.9\%}} \\
\hline

0.00
& 59.41 & 59.36 & 1626 & 86.72 & 68.52 & 55.65 & 95.0\% \\

0.15 
& 58.56 & 59.97 & 1650 & 86.82 & 69.60 & 56.16 & 95.6\% \\

0.30 
& 58.63 & 60.14 & 1650 & 86.94 & 69.75 & 56.13 & 95.7\% \\

0.45 
& \textbf{59.12} & \textbf{60.58} & \textbf{1665} & \textbf{86.98} & \textbf{69.80} & \textbf{56.32} & \textcolor{red!80!black}{\textbf{96.2\%}} \\

0.60 
& 58.65 & 60.05 & 1644 & 86.96 & 69.80 & 56.09 & 95.7\% \\
\hline

\specialrule{0.8pt}{0pt}{0pt}
\end{tabular}

{\fontsize{9}{10}\selectfont
\caption{Ablation study of $\eta$ on LLaVA-1.5-7B.}
\label{tab:supplement_eta_ablation}
}

\end{table}

\paragraph{Impact of $\eta$, $\theta_s$ and $\theta_c$.}
After PRCA identifies the calibration anchors, CGTC calibrates them through a two-stage procedure. The first stage applies a confidence gate to select reliable calibration signals from the residual, unselected tokens. This stage is controlled by three hyperparameters: the
spatial-proximity weight $\eta$ in the semantic-spatial affinity, the best-affinity threshold $\theta_s$, and the confidence threshold $\theta_c$. The coefficient $\eta$ balances semantic similarity and spatial proximity when matching a residual token to an anchor. Semantically similar tokens often correspond to the same object or region, but visually similar features may also arise from different objects. Spatial proximity is therefore incorporated to reduce
incorrect cross-object assignments. In practice, an anchor tends to absorb semantically compatible signals from its local neighborhood, thereby enriching its representation without introducing unrelated content. Based on the Table \ref{tab:supplement_eta_ablation}, we set $\eta=0.45$.

The thresholds $\theta_s$ and $\theta_c$ jointly control signal admission. In Figure~\ref{figure8}, fixing $\theta_s=0.4$ and increasing $\theta_c$ through 0.30, 0.45, 0.60, 0.75, and 0.90 reduces acceptance from 100.0\% to 99.7\%, 78.3\%, 55.7\%, and 39.2\%, respectively. The corresponding F1 scores are 82.14, 82.23, 83.85, 84.96, and 85.66. Thus, admitting nearly all residual tokens is less effective in this grid than filtering them before calibration. At $\theta_c=0.90$, the three affinity thresholds $\theta_s\in\{0.3,0.4,0.5\}$ yield similar F1 scores of 85.65, 85.66, and 85.65. Further increasing $\theta_s$ to 0.7 reduces acceptance to 33.3\% and F1 to 85.31, suggesting that excessive filtering also removes useful complementary signals. We select $(\theta_s,\theta_c)=(0.4,0.9)$, which gives the highest displayed F1 while retaining 39.2\% of residual tokens as calibration signals. This comparison identifies a favorable tested configuration, rather than establishing a universal or statistically significant optimum.

\paragraph{Impact of $\alpha$.}
In the second stage of CGTC, the selected signals are fused into their matched calibration anchors. The coefficient $\alpha$ controls the strength of this residual injection. Calibration signals are intended to complement and reinforce the anchor representation rather than dominate it. A small $\alpha$ prevents useful residual semantics from being effectively transferred, whereas an excessively large value may overwrite the original anchor representation and make the procedure resemble conventional token merging. According to the Table \ref{tab:supplement_alpha_ablation}, we set $\alpha=0.15$. Overall, the ablation studies demonstrate that RefineV remains robust across a broad range of hyperparameter settings.

\begin{table}[t]
\centering
\fontsize{9}{11}\selectfont
\setlength{\tabcolsep}{4pt}
\renewcommand{\arraystretch}{1.11}

\begin{tabular}{c|cccccc|c}
\specialrule{0.8pt}{0pt}{0pt}
\textbf{$\alpha$} 
& GQA & MMB & MME & POPE & SQA & VQA$^T$ & Avg. \\
\specialrule{0.8pt}{0pt}{0pt}

\rowcolor{gray!20}
\multicolumn{8}{c}{\textit{Total 576 Tokens}} \\
\hline
-
& 61.90 & 64.70 & 1862 & 85.90 & 69.50 & 58.20 & 100\% \\
\hline

\rowcolor{gray!20}
\multicolumn{8}{c}{\textit{Retain 192 Tokens} \textcolor{green!70!black}{$\downarrow$ 66.7\%}} \\
\hline

0.00
& 60.11 & 63.01 & 1784 & 87.78 & 69.80 & 57.19 & 98.5\% \\

0.10 
& 60.20 & 62.72 & 1792 & 88.70 & 69.83 & 57.87 & 98.9\% \\

0.15 
& \textbf{60.57} & \textbf{63.41} & \textbf{1809} & \textbf{88.94} & \textbf{70.24} & \textbf{58.01} & \textcolor{red!80!black}{\textbf{99.6\%}} \\

0.20 
& 60.45 & 63.36 & 1789 & 88.60 & 69.86 & 57.81 & 99.1\% \\

\hline

\rowcolor{gray!20}
\multicolumn{8}{c}{\textit{Retain 128 Tokens} \textcolor{green!70!black}{$\downarrow$ 77.8\%}} \\
\hline

0.00
& 59.72 & 61.36 & 1733 & 87.50 & 69.42 & 56.62 & 97.2\% \\

0.10 
& 59.63 & 61.86 & 1743 & 87.51 & 69.06 & 56.99 & 97.5\% \\

0.15 
& \textbf{60.15} & \textbf{62.11} & \textbf{1731} & \textbf{87.79} & \textbf{69.89} & \textbf{57.96} & \textcolor{red!80!black}{\textbf{98.1\%}} \\

0.20 
& 59.83 & 62.11 & 1724 & 87.50 & 69.42 & 56.62 & 97.4\% \\
\hline

\rowcolor{gray!20}
\multicolumn{8}{c}{\textit{Retain 64 Tokens} \textcolor{green!70!black}{$\downarrow$ 88.9\%}} \\
\hline

0.00
& 58.61 & 60.23 & 1632 & 86.28 & 68.71 & 56.26 & 95.2\% \\

0.10 
& 58.47 & \textbf{60.65} & 1626 & 86.56 & 68.67 & 56.14 & 95.3\% \\

0.15 
& \textbf{59.12} & 60.58 & \textbf{1665} & \textbf{86.98} & \textbf{69.80} & \textbf{56.32} & \textcolor{red!80!black}{\textbf{96.2\%}} \\

0.20 
& 58.89 & 60.57 & 1654 & 86.59 & 68.67 & 56.33 & 95.7\% \\
\hline

\specialrule{0.8pt}{0pt}{0pt}
\end{tabular}

{\fontsize{9}{12}\selectfont
\caption{Ablation study of $\alpha$ on LLaVA-1.5-7B.}
\label{tab:supplement_alpha_ablation}
}

\end{table}

\section{E. Additional Qualitative Results}
\subsection{More Visualization Examples}
\begin{figure*}[t]
\centering
\includegraphics[scale=0.7]{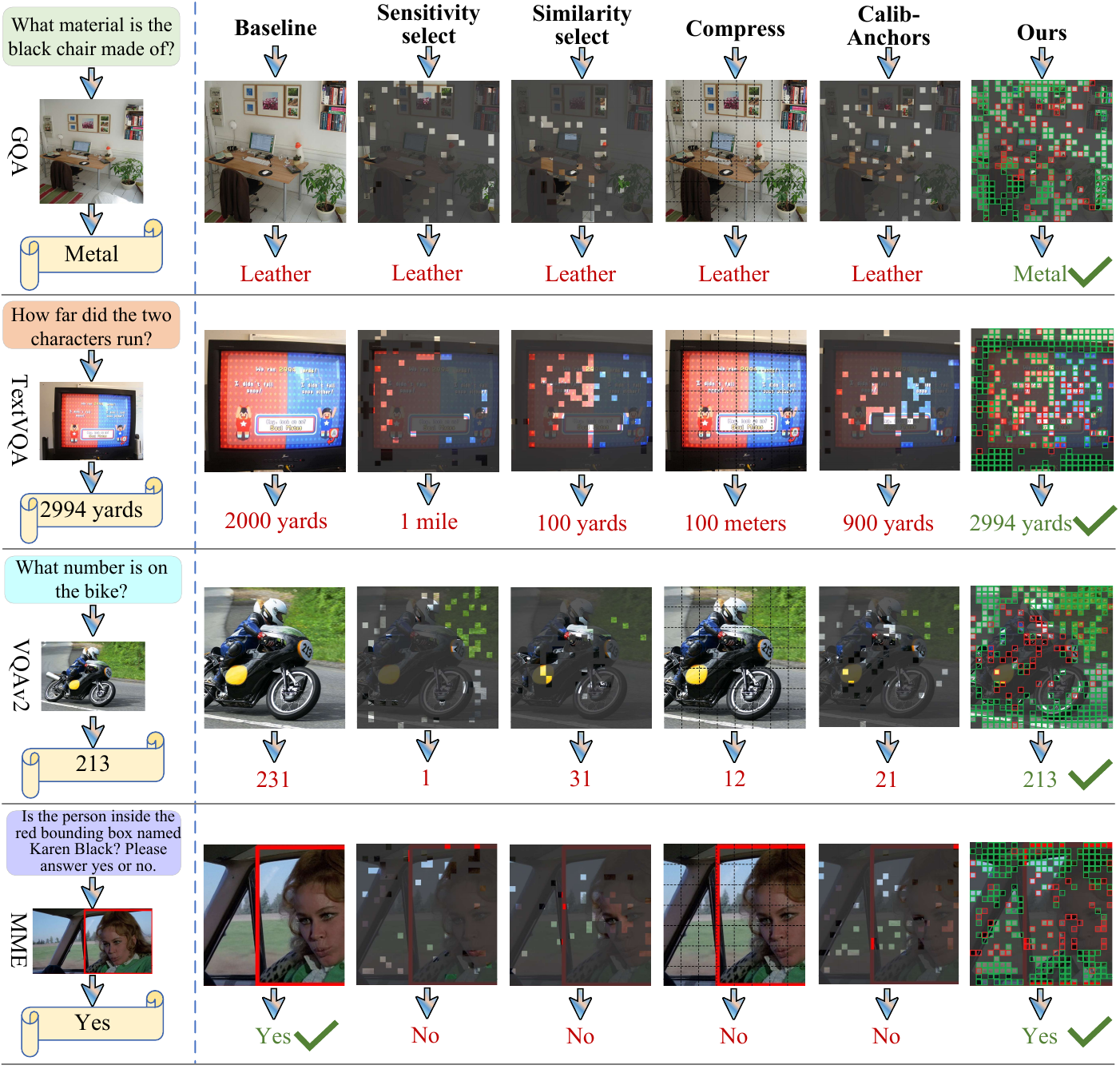} 
\caption{Qualitative visualizations on LLaVA-1.5-7B across four tasks, including GQA, VQA$^T$, VQA$^{v2}$ and MME. Pruning ratio is 88.9\%. }
\label{figure_visual_llava_1.5}
\end{figure*}

Figures \ref{figure_visual_llava_1.5}, \ref{figure_visual_llava_next} and \ref{figure_visual_qwen} provide additional qualitative comparisons on LLaVA-1.5-7B, LLaVA-NeXT-7B and Qwen2.5-VL-7B, respectively. For each backbone, we visualize representative examples from four tasks, including GQA, $\mathrm{VQA}^{\mathrm{T}}$, $\mathrm{VQA}^{\mathrm{v2}}$ and MME. The leftmost part of each example presents the original question, input image, and ground-truth answer. The remaining columns show the detailed token-processing results and all model predictions produced by six different configurations.

``\textbf{Baseline}'' denotes the original model that performs inference using all visual tokens without token reduction. ``\textbf{Sensitivity Select}'' retains tokens solely according to their sensitivity scores, whereas ``\textbf{Similarity Select}'' performs token selection based on semantic similarity. These two settings represent conventional score-based token selection strategies. ``\textbf{Compress}'' represents feature-compression methods that aggregate the original visual tokens into a fixed number of compact tokens before subsequent inference. ``\textbf{Calib-Anchors}'' corresponds to an incomplete variant of our framework, in which only perturbation-robust calibration anchors are retained for inference without incorporating additional calibration signals. Finally, ``\textbf{Ours}'' denotes the complete RefineV framework. It first identifies stable calibration anchors and then selects informative tokens from the remaining token set as calibration signals, which are used to refine the semantic representations of the anchor tokens.

In the visualizations of RefineV, tokens enclosed by red boxes denote calibration anchors, while those enclosed by green boxes represent the selected calibration signals. Tokens without either annotation are discarded. This decomposition illustrates that RefineV does not indiscriminately retain or merge visual information. Instead, it preserves a compact set of semantically stable anchors and selectively exploits complementary evidence from the remaining tokens to correct and enrich their representations.

The qualitative examples illustrate the limitations of conventional token reduction. Sensitivity- and similarity-based selection can overlook sparse but decisive evidence, such as object materials, scene text, numerical identifiers, colors, spatial relations and clock readings. Feature compression may retain broader image coverage, but aggregating heterogeneous regions can dilute fine-grained semantics and produce incorrect answers. These failures indicate that simply retaining high-scoring tokens or merging nearby features does not guarantee preservation of task-relevant semantics.

A comparison between ``\textbf{Calib-Anchors}'' and ``\textbf{Ours}'' highlights the importance of calibration. While robust anchors retain salient regions, using them alone can still cause semantic deviations. By incorporating reliable calibration signals, RefineV preserves and reassigns complementary evidence and corrects these errors, sometimes even outperforming the unpruned baseline. These results support modeling token reduction as token calibration and show its effectiveness in mitigating semantic drift.

\subsection{Failure Case Examples}
\begin{figure*}[t]
\centering
\includegraphics[scale=0.7]{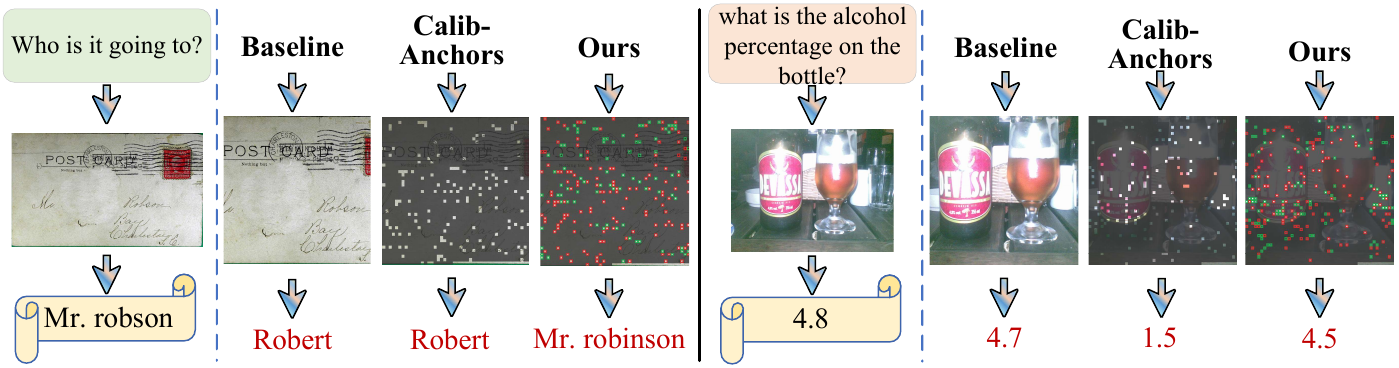} 
\caption{Failure cases of RefineV at a 94.4\% pruning ratio. }
\label{figure_failure_case}
\end{figure*}

\begin{figure*}[t]
\centering
\includegraphics[scale=0.7]{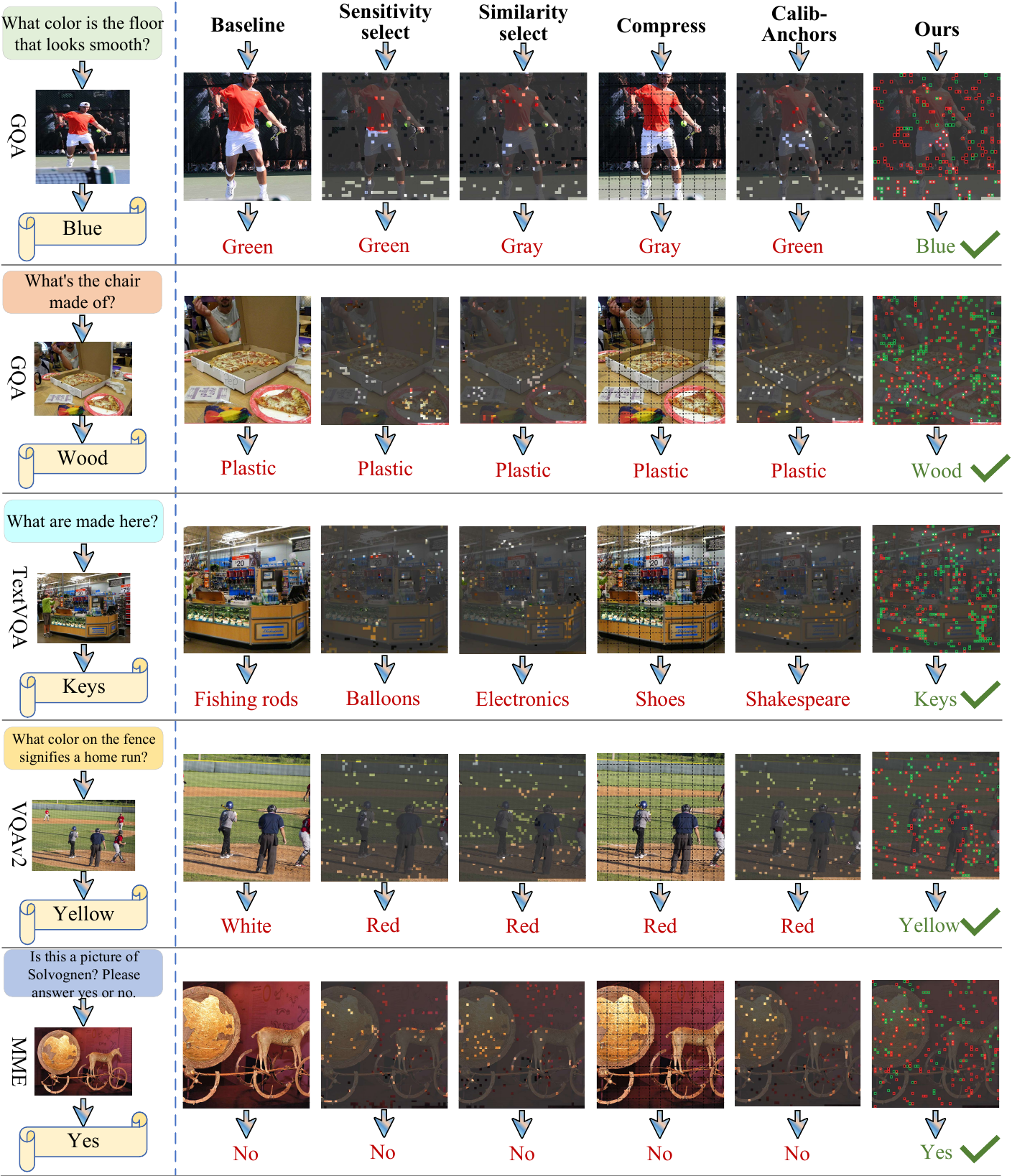} 
\caption{Qualitative visualizations on LLaVA-NeXT-7B across four tasks, including GQA, VQA$^T$, VQA$^{v2}$ and MME. Pruning ratio is 94.4\%. }
\label{figure_visual_llava_next}
\end{figure*}

\begin{figure*}[t]
\centering
\includegraphics[scale=0.7]{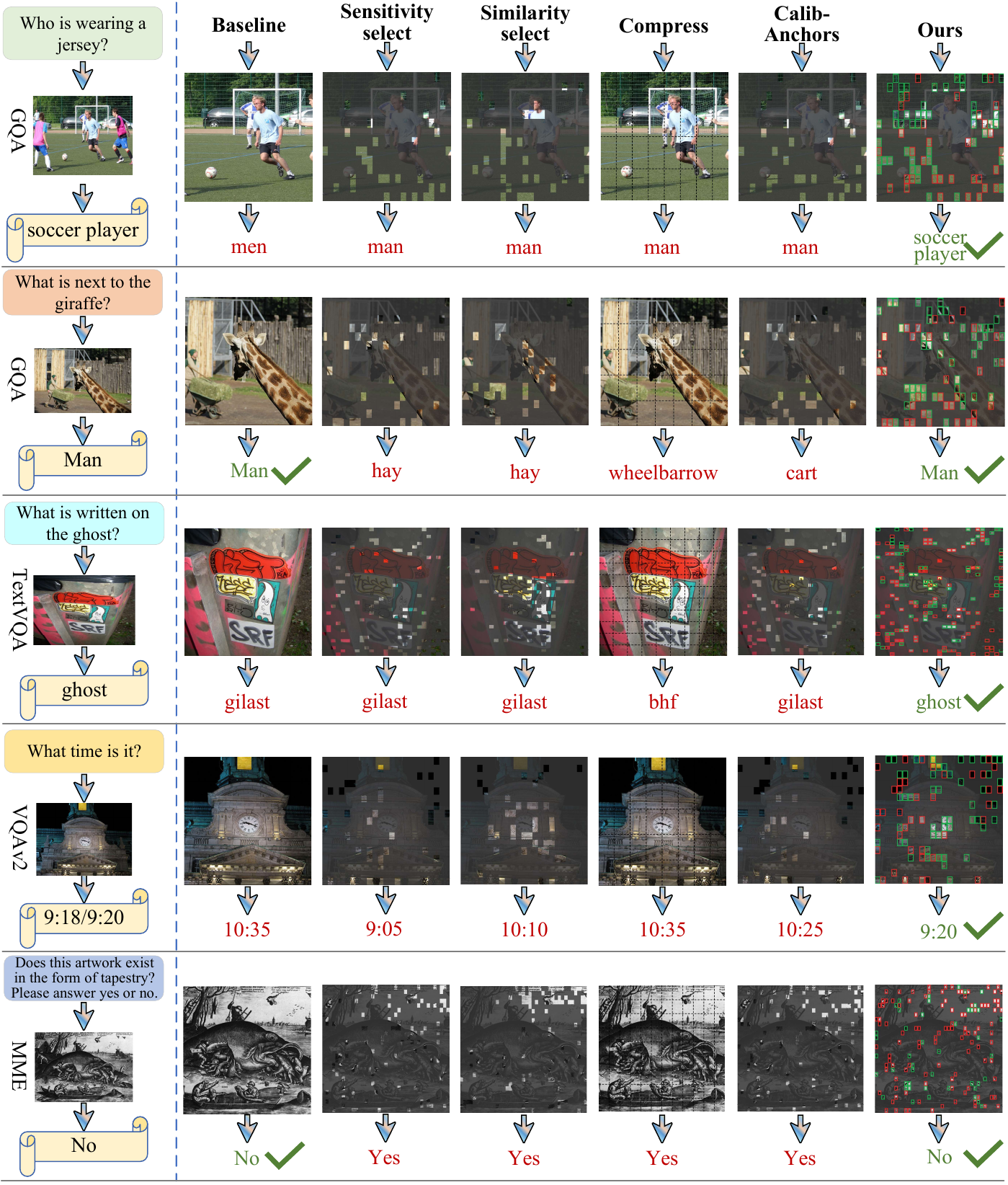} 
\caption{Qualitative visualizations on Qwen2.5-VL-7B across four tasks, including GQA, VQA$^T$, VQA$^{v2}$ and MME. Pruning ratio is 90\%. }
\label{figure_visual_qwen}
\end{figure*}

As shown in Figure \ref{figure_failure_case}, we further present several representative failure cases. These examples reveal a limitation in fine-grained visual grounding under ambiguous or low-resolution textual cues. Both require exact transcription of names or numbers rather than coarse semantic recognition. Although RefineV preserves task-relevant evidence, calibration cannot reconstruct information that is weakly encoded, visually corrupted, or inconsistently represented across tokens; language priors may then dominate decoding. Thus, performance remains sensitive to OCR fidelity and token-level evidence quality, reflecting a deeper information bottleneck rather than a simple selection error.
\section{F. Discussion and Future Work}
RefineV reframes fixed-budget visual token reduction as a calibration problem rather than a purely selective one, suggesting that preserving semantic fidelity requires not only retaining informative tokens but also refining their representations before multimodal reasoning. Its training-free and backbone-agnostic formulation makes this principle broadly applicable to existing reduction pipelines, where discarded tokens can serve as auxiliary evidence for calibrating robust anchors. Nevertheless, RefineV remains constrained by the quality of the original visual representations. When critical evidence is weakly encoded, highly ambiguous, or absent from the token sequence, projector-level calibration cannot fully reconstruct the missing information. In our view, additional directions include adaptive calibration under varying token budgets and uncertainty-aware signal selection, which may further improve semantic completeness across heterogeneous architectures and multimodal inputs.

